\documentclass[final,1p,times]{elsarticle}

\usepackage{amssymb}
\usepackage{amsthm}
\usepackage{amsmath}
\usepackage{tikz-cd}

\usepackage{enumitem}

\newtheorem{theorem}{Theorem}
\newtheorem{proposition}{Proposition}
\newtheorem{corollary}{Corollary}
\newtheorem{lemma}{Lemma}

\theoremstyle{definition}
\newtheorem{definition}{Definition}

\theoremstyle{definition}

\theoremstyle{definition}
\newtheorem{example}{Example}

\journal{International Journal of Approximate Reasoning}

\begin{document}

\begin{frontmatter}



\title{Projective families of distributions revisited}


\author{Felix Weitk\"amper}
\ead{felix.weitkaemper@lmu.de}

\address{Institut f\"ur Informatik, Ludwig-Maximilians-Universit\"at M\"unchen,
             Oettingenstr.~67, 
             80538 M\"unchen,
             Germany}

\begin{abstract}
The behaviour of statistical relational representations across differently sized domains has become a focal area of research from both a modelling and a complexity viewpoint. 
  Recently, projectivity of a family of distributions emerged as a key property, ensuring that marginal probabilities are independent of the domain size.
  However, the formalisation used currently assumes that the domain is characterised only by its size. This contribution extends the notion of projectivity from families of distributions indexed by domain size to functors taking extensional data from a database. This makes projectivity available for the large range of applications taking structured input. 
  We transfer key known results on projective families of distributions to the new setting.
  This includes a characterisation of projective fragments in different statistical relational formalisms as well as a general representation theorem for projective families of distributions. 
  Furthermore, we prove a correspondence between projectivity and distributions on countably infinite domains, which we use to unify and generalise earlier work on statistical relational representations in infinite domains. 
  Finally, we use the extended notion of projectivity to define a further strengthening, which we call $\sigma$-projectivity, and which allows the use of the same representation in different modes while retaining projectivity.
\end{abstract}

\begin{keyword}
 Infinite domains \sep Projectivity \sep Structured model \sep Statistical relational artificial intelligence \sep Lifted probabilistic inference
\end{keyword}

\end{frontmatter}


\section{Introduction}
Statistical relational artificial intelligence (AI) comprises approaches that combine probabilistic learning and
reasoning  with variants of first-order predicate logic.
The challenges of statistical relational AI have been adressed from both directions:
Either probabilistic graphical models such as Bayesian networks or Markov networks are
lifted to relational representations and linked to (variants of) first-order logic, or
approaches based on predicate logic such as logic programming
are extended to include probabilistic facts.
The resulting statistical relational languages make it possible to specify a
complex probabilistic model compactly and without reference to a specific domain of objects.

Formally, on a given input, a statistical relational model defines a probability distriution over possible worlds on the domain of the input, which can then be queried for the probabilites of various definable events. 

Compared to ordinary Bayesian networks or Markov networks, statistical relational AI offers
several advantages:

\begin{itemize}
  \item The presentation is generic, which means that it can be transferred to other areas with a similar structure

  \item It is  possible to specify complex background knowledge declaratively.
For example, different modelling assumptions can be implemented and adapted rapidly.

  \item Statistical relational approaches allow probabilistic and logical inference query tasks such as
abductive and deductive inference to be combined seamlessly.

  \item Known symmetries can be enforced when learning the structure or the parameters of the model --
this makes it possible to smooth out known random fluctuations in the data set and achieve more coherent models.

  \item Finally, compact and domain-independent models are easy for humans to read and check for
    plausibility. In this way statistical relational AI contributes significantly to the search for powerful explainable AI models.
\end{itemize}

The compact and domain-independent representation of a statistical relational model is one of its
main advantages. Therefore, one expects the model to behave intuitively when applied to object
domains of different sizes.
However, this is generally not the case with any of the above approaches.
To the contrary, it is the rule rather than the exception that the limits of the probabilities of statements are completely independent of the parameters of the model as the domain size increases \citep{PooleBKKN14}.

The biggest practical challenge of statistical relational AI, however, is the scalability of learning
and inference on larger domains. While various approaches have been developed in the last
decade that take advantage of the unified specification to solve inference tasks without actually
instantiating the network on the given domain, they are restricted by the inherent complexity of
the  task: Inference in typical specification languages is \#P-hard in the size of the domain \citep{BeameBGS15}. 
This is even more painfully felt in learning, as many inference queries are usually executed during a single learning process.

These observations suggest the concept of a \emph{projective family of distributions}.
Essentially, a family of probability distributions defined on different domains is projective if the response to queries referring to elements of a smaller subdomain does not depend on the size of the entire domain.
  \begin{example} \label{SBM}
A typical example of a projective family of distributions is the relational stochastic block model \citep{HollandLL83,MalhotraS22} with two communities $C_0$ and $C_1$, a probability $P$ of a given node to lie in community $C_1$, and edge probabilities $p_{ij}$ between nodes of communities $C_i$ and $C_j$. In this model, all choices of community are made independently in a first step and then the choices of edge existence are made independently of each other with the probabilities corresponding to the communities of the two nodes.
  \end{example}
In projective families, marginal inference is possible without even considering the domain
itself, or its size. Thus, the marginal inference problem can be solved in time depending only on the query, regardless of domain size.

  Statistical relational frameworks are well established as a method for probabilistic learning and reasoning in highly structured domains.
  They are used in a variety of ways, from full generative modelling to prediction tasks from data. 
  Many applications  lie between those extremes, taking structured extensional data as input and providing a generative model of the intensional vocabulary as output.
  \begin{example}\label{lead}
      Consider the following example domains for network-based models:
  \begin{enumerate}[label=\alph*]
      \item\label{randgraph} A typical application domain of full generative models are random graph models, which provide a declarative specification for generating random graphs, potentially with some extra structure.
  An example is the relational stochastic block model from Example \ref{SBM} above. 
  
  \item On the other end of the spectrum are link prediction tasks \citep{LuZ11}; here, the nodes, the colouring if applicable, and a subset of edges are provided as input. The task is to predict the existence of the missing edges. 
  
  \item\label{multilink} As a typical example in between those extremes, consider link prediction over multiple networks \citep{AhmadBSC10}, where a range of prior knowledge about the individuals is considered, including node attributes and connections from other networks.

  \item\label{epidemiology} Network-based epidemiological modelling \citep{DanonFHJKRRV11} is another active application domain of a mixed type. Here, the output is a generative model of the spread of a disease, while an underlying contact network is given as data. 
  \end{enumerate}
  \end{example}
  
  In the statistics literature, projectivity was explored by Shalizi and Rinaldo in the context of random graph models  \cite{ShaliziR13}. Jaeger and Schulte  then extended the notion to general families of distributions defined by a variety of statistical relational formalisms \cite{JaegerS18}. Later, they gave a complete characterisation of projective families of distributions in terms of random arrays \cite{JaegerS20} .

  Jaeger and Schulte  also demonstrated the projectivity of certain limited syntactic fragments of probabilistic logic programming, relational Bayesian networks and Markov logic networks \cite{JaegerS18}.
On the other hand, it has been demonstrated that common statistical relational formalisms such as probabilistic logic programs and 2-variable Markov logic networks can only express a very limited fragment of this rich class of families \citep{Weitkaemper21,MalhotraS22}. 

 This body of research  assumes that the domain is characterised only by its size and can therefore be presented as an initial segment of the natural numbers. This restricts the concept to applications of the type of Example \ref{lead}.\ref{randgraph}.

 In a situation of richer input data, taken from an extensional database, it is natural to see a model not just as an indexed family of distributions, but as a map that takes structures in the extensional vocabulary as input.

We generalise the concept of projectivity to this setting and show that the main results from \cite{JaegerS18,JaegerS20,Weitkaemper21} carry over. In particular, we introduce AHK representations for structured input and prove an analogue of the representation theorem in \cite{JaegerS20}.

We also demonstrate a one-to-one correspondence between projective families of distributions and exchangeable distributions on a countably infinite domain.
This relates the present line of work to earlier results on infinite structures
and can streamline the results in that area.
We then generalise this correspondence to structured input, suggesting projectivity as an interesting framework for probabilistic reasoning over dynamic models and data streams.
Finally, we introduce $\sigma$-projective families of distributions, which remain projective even when conditioning on a subvocabulary, and apply this notion to obtain an inexpressivity result for the $\sigma$-determinate Markov logic networks introduced by Singla and Domingos \cite{SinglaD07}.

\section{Preliminaries}

We introduce the concepts and notation from logic, probability and statistical relational artificial intelligence referred to in this paper.

\subsection{Logical preliminaries}
We begin with the logical syntax:
A \emph{vocabulary} $L$ consists of a set of \emph{relation symbols} $R$ with a given \emph{arity} $m_R$, and a set of constants $c$. 
$L$ is \emph{relational} if it is does not contain any constants.
An \emph{$L$-atom} is an expression of the form $R(x_1, \dots,x_n)$, where $x_1, \dots, x_n$  are either constants from $L$ or from a countably infinite set of \emph{variables} that we assume to be available. 
Additionally, expressions of the form $x_1 = x_2$ are considered atoms.
An \emph{$L$-literal} is either an atom or an expression of the form $\neg \varphi$, where $\varphi$ is an atom.
A \emph{quantifier-free $L$-formula} is built up recursively from $L$-atoms using the unary connective $\neg$ and the binary connectives $\wedge$ and $\vee$. 
A quantifier-free $L$-formula is called a \emph{sentence} if it contains no variables.

The semantics is defined by \emph{$L$-structures}:
Let $D$ be a set. Then an \emph{$L$-structure $\mathfrak{X}$ on domain $D$} is an interpretation of $L$, that is, for every relation symbol $R$ of arity $m$ in $L$ a subset $R^{\mathfrak{X}}$ of $D^m$, and for every constant $c$ in $L$ an element $c^{\mathfrak{X}}$ of $D$. 

An {\em embedding of $L$-structures} from $\mathfrak{X}_1$ on domain $D_1$ is to $\mathfrak{X}_2$ on domain $D_2$ is an injective map $\iota$ from $D_1$ to $D_2$ such that for any constant $c$, the interpretation of $c$ in $D_1$ is mapped to the interpretation of $c$ in $D_2$ and for any relation symbol $R$ of arity $m$ and for any $m$-tuple $(a_1, \ldots, a_m)$ in $D_1$, $(a_1, \ldots, a_m)$ lies in the interpretation of $R$ in $D_1$ if and only if $(\iota(a_1), \ldots \iota(a_m))$ lies in the interpretation of $R$ in $D_2$. A bijective embedding is an {\em isomorphism of $L$-structures}, or an {\em automorphism} if domain and co-domain coincide. 

If $D$ is a set, $L_D$ denotes the language $L$ enriched by constants $c_a$ for every element $a \in D $. We call a quantifier-free $L_D$-sentence a quantifier-free $L$-\emph{query} over $D$.
A formula is \emph{grounded }by substituting elements of $D$ for its variables, and it is \emph{ground
}if it does not (any longer) contain variables.
Therefore, any choice
of elements of $D$ matching the variables in
a formula is a \emph{possible grounding }of that formula.

An $L$-structure $\mathfrak{X}$ \emph{models} a ground quantifier-free $L$-formula $\varphi$ if $\varphi$ is true for the interpretations in $\mathfrak{X}$, where the connectives $\neg$, $\wedge$ and $\vee$ are interpreted as `not', `and' and `or' respectively.
A quantifier-free formula $\varphi$ is \emph{consistent} if there is a set $D$, an $L$-structure $\mathfrak{X}$ with domain $D$ and a grounding of $\varphi$ that is modelled by $\mathfrak{X}$.
A quanitifer-free formula is consistent \emph{with} another quantifier-free formula if their conjunction is consistent.

For any quantifier-free formula $\varphi (x_1, \dots, x_n)$ with variables from $x_1, \dots, x_n$, we denote by $\varphi (a_1, \dots, a_n)$ for $a_1, \dots,a_n \in D$ the  quantifier-free $L$-query over $D$ obtained by substituting $c_{a_i}$ for $x_i$.  
It is easy to see that every finite structure $\mathfrak{X}$ on $\{a_1, \dots,a_n\}$ can be uniquely described by a quantifier-free $L$-query over $\{a_1, \dots,a_n\}$.
We refer to the formula $\varphi(x_1, \dots,x_n)$ for which $\varphi(a_1, \dots,a_n)$ uniquely describes $\mathfrak{X}$ as the \emph{$L$-type} of $\mathfrak{X}$, and we call the $\emptyset$-type the \emph{$=$-type} to emphasise that $=$ can be used in atoms even if $L = \emptyset$. If $L$ is clear from context, we will also write \emph{$n$-type} to emphasise the arity. Every type can be canonically expressed as a conjunct of distinct literals. 
It will be occasionally convenient to subdivide the $L$-type $\varphi$ further; call the conjunction of those literals containing exactly the variables $x_{i_1}, \dots, x_{i_m}$ and without the equality sign the \emph{data of arity m of $(a_{i_1}, \dots, a_{i_m})$}, denoted $\varphi^m$. Up to logical equivalence, there are only finitely many types with the same set of variables. We call this finite set $\mathcal{T}^{L}$, and the set of all possible data of arity $m$, $\mathcal{T}_m^{L}$.

Injective maps $\iota: D' \hookrightarrow D$ between sets induce a natural map from $L$-structures $\mathfrak{X}$ on $D'$ to $L$-structures $\iota (\mathfrak{X})$ on the image set $\iota (D')$:
Simply interpret $R$ by the set
$\{(\iota(a_1), \dots, \iota(a_m))\mid a_1, \dots, a_m \in R^{\mathfrak{X}}\}$,
and set
$c^{\iota (\mathfrak{X})} := c^{\mathfrak{X}}$. 

Let $D' \subseteq D$. Then we call an $L$-structure $\mathfrak{Y}$ on $D$ an \emph{extension} of an $L$-structure  $\mathfrak{X}$ on $D'$ if $R^{\mathfrak{Y}} \cap {(D')}^m = R^{\mathfrak{X}}$ for every relation symbol $R$ in $L$ and $c^{\mathfrak{Y}} = c^{\mathfrak{X}}$ for every constant symbol $c$.
$\mathfrak{X}$ is then also called the $L$-substructure of $\mathfrak{Y}$ on $D$. 

On the other hand, consider vocabularies $L' \subseteq L$, a set $D$, an $L'$-structure $\mathfrak{X}$ and an $L$-structure $\mathfrak{Y}$. Then  $\mathfrak{Y}$ is called an \emph{expansion} of  $\mathfrak{X}$ if the interpretations of the symbols of $L'$ coincide in  $\mathfrak{X}$ and  $\mathfrak{Y}$, and we write $\mathfrak{Y}_{L'}$ for $\mathfrak{X}$.

\begin{example}\label{logprelimexample}
  We illustrate some of these notions using the example of coloured graphs.
  Consider a signature $L$ with a binary edge relation $E$ and a unary relation $P$.
  Then an $L$-structure $G$ is a directed graph with edge relation $E$, on which $P$  divides the nodes into two disjoint sets (those $a\in G$ for which $P(a)$ holds and those for which $P(a)$ does not hold).

  Quantifier-free $L$-queries are those which ask whether a specific node has a certain colour, or whether a specific pair of nodes is connected by an edge, or Boolean combinations thereof;
  a query as to whether \emph{any} two nodes are connected by an edge cannot be expressed with a quantifier-free $L$-query.

  A 1-type in this signature specifies which colour a node has, and whether the node has a loop.
  A 2-type specifies the 1-types of a given pair of nodes $(a,b)$, whether there are edges from $a$ to $b$ and/or vice versa. This additional information is the data of arity 2.

  If $H$ is a coloured subgraph of $G$, then $G$ is an extension of $H$; if $G'$ is the underlying uncoloured graph of $G$, then $G$ is an expansion of $G'$. 
\end{example}

\subsection{Probabilistic preliminaries}
As we are interested in probabilistic models, we introduce the terminology that we adopt for decribing probabilistic models.
For every finite set $D$ and vocabulary $L$, let $\Omega^{D}_L$ be the set of all $L$-structures on the domain $D$. 
We consider probability distributions $P$ defined on the power set of (the finite set) $\Omega^{D}_L$, and call them \emph{$L$-distributions over $D$},
where $L$ is omitted if it is clear from context.
$P$ is completely defined by its value on the singleton sets $P(\{\mathfrak{X}\})$, and we write $P(\mathfrak{X})$ for $P(\{\mathfrak{X}\})$. 
As elements of  $\Omega^{D}_L$, $L$-structures are also known as \emph{possible worlds}.
In this context, subsets of the probability space  $\Omega^{D}_L$ are known as \emph{events}, and we
frequently write
$P(\text{a property of } \mathfrak{Y})$ for
$P(\{\mathfrak{Y}\mid \text{a property of }  \mathfrak{Y}\})$
where the set comprehension variable is by convention the first variable to appear in the statement of the property. 
So, for instance, $P(\mathfrak{Y}\text{ extends }\mathfrak{X})$ stands for $P(\{\mathfrak{Y} \in \Omega^{D}_L\mid \mathfrak{Y}\text{ extends }\mathfrak{X}\})$.
This also allows us to write \emph{conditional probablities}, where
\[P(\text{First property of } \mathfrak{Y}\mid \text{Second property of } \mathfrak{Y})\]
stands for \[P(\text{First and second property of } \mathfrak{Y})\div P(\text{Second property of } \mathfrak{Y}),\]
which is well-defined whenever the probability of the second property is positive.
When $\varphi$ is a query over a finite set $D$ and $P$ a distribution over $D$, then we call $P(\{\mathfrak{X} \in \Omega^{D}_L \mid \mathfrak{X} \models \varphi\})$ the \emph{marginal probability of $\varphi$ under P}, which we write simply as $P(\varphi)$. 

An \emph{$L$ family of distributions} is a map taking a finite set as input and returning an $L$-distribution over $D$. 
When discussing the notion of projectivity from \cite{JaegerS18,JaegerS20}, we also refer to \emph{$\mathbb{N}$-indexed $L$ families of distributions}, which only take initial segments of $\mathbb{N}$ as input. 
In this case, the distribution over $\{1, \dots, n\}$ is denoted $P_n$ in line with \cite{JaegerS18,JaegerS20}.
We use the shorthand notation $(P)$ for an ($\mathbb{N}$-indexed) $L$ family of distributions ${(P_D)}_{D\textrm{ a finite set}}$ (resp. ${(P_n)}_{n \in \mathbb{N}}$).

\begin{example}
  Continuing the example of coloured graphs, let $D$ be a set, and let $L = \{E,P\}$ for a binary $E$ and a unary $P$. Then $\Omega^D_L$ is the set of all coloured graphs on the node set $D$.
  An $L$ family of distributions would allocate every finite node set $D$ a probability distribution on the finite set $\Omega^D_L$, while an $\mathbb{N}$-indexed $L$ family of distributions would do the same, but only take node sets of the form $\{1,\dots,n\}$ as input. 
\end{example}

\subsection{Statistical relational artificial intelligence}\label{StarAI}
Over the past 30 years, a variety of different formalisms have been suggested for combining relational logic with probabilities. Here we outline and analyse three of those formalisms, which exemplify different strands within statistical relational artificial intelligence:\@
Relational Bayesian Networks (RBN), introduced by Jaeger \cite{Jaeger97}, lift Bayesian networks to relationally structured domains;
Markov Logic Networks (MLN), introduced by Richardson and Domingos \cite{RichardsonD06}, are based on undirected Markov networks rather than on directed Bayesian networks;
Probabilistic logic programs (PLP) in form of ProbLog programs, introduced by De Raedt and Kimmig \cite{deRaedtK07} but based on the distribution semantics introduced earlier by Sato \cite{Sato95}, add probabilistic primitives to logic programming.
We only give a brief account of each of the formalisms here and refer the reader to the cited literature for more details.
We start with RBNs:
\begin{definition}
  An \emph{$L$-probability formula} with free variables $\mathrm{fv}$ is inductively defined as follows:
  \begin{enumerate}
  \item Each $q \in [0,1]$ is a probability formula with $\mathrm{fv}(q)=\emptyset$.
  \item For each $R \in L$ of arity $m$ and variables $x_1, \dots, x_m$, $R(x_1, \dots, x_m)$ is a probability formula with $\mathrm{fv}(R(x_1, \dots, x_m))=\{x_1, \dots, x_m\}$.
  \item When $F_1$, $F_2$ and $F_3$ are probability formulas, then so is $F_1 \cdot F_2 + (1 - F_1) \cdot F_3$ with $\mathrm{fv}(F_1 \cdot F_2 + (1 - F_1) \cdot F_3) = \mathrm{fv}(F_1)\cup \mathrm{fv}(F_2) \cup \mathrm{fv}(F_3)$.
  \item When $F_1, \dots, F_k$ are probability formulas, $\vec{w}$ is a tuple of variables and $\mathrm{comb}$ a function that maps finite multisets with elements from $[0,1]$ into $[0,1]$, then $\mathrm{comb}(F_1, \dots, F_k \mid \vec{w})$ is a probability formula with $\mathrm{fv}(\mathrm{comb}(F_1, \dots, F_k \mid \vec{w})) = \mathrm{fv}(F_1, \dots, F_k) \setminus \vec{w}$.
  \end{enumerate}

  A \emph{Relational Bayesian Network} (with vocabulary $L$) is an assignment of $L$-probability formulas $F_R$ to relation symbols $R$ along with $\mathrm{arity}(R)$ many variables $x_1, \dots, x_m$, such that $\mathrm{fv}(F_R)\subseteq \{x_1, \dots, x_m\}$ and such that the dependency relation $S \leq R$, which holds whenever $S$ occurs in $F_R$, is acyclic. $F_R$ is called the \emph{label} of $R$. 
\end{definition}

\begin{example}\label{Ex_RBN}
  Consider a vocabulary $L = \{R,S\}$ of two unary relation symbols.
  Then the probability formulas $F_R = 0.7\cdot S(x) + 0.2\cdot (1 - S(x))$ with free variable $x$ and probability formula $F_S = 0.5$ define an RBN  $B_1$ without combination functions.
  The probability formulas $F_R = \mathrm{arithmetic mean}(S(y)\mid y)$ and $F_S = 0.5$ define an RBN $B_2$ with a combination function.
\end{example}

\begin{definition}
  The semantics of an RBN is given by grounding to a Bayesian network. Let $D$ be a finite set.
  For every query atom $R(a_1, \dots, a_m)$, obtain $F_{R(a_1, \dots, a_m)}$ from $F_R$ by substituting $a_1, \dots, a_m$ for the free variables $x_1, \dots, x_m$ respectively.
  Consider the directed acyclic graph $G$ whose nodes are query atoms over $D$.
  Draw an edge between nodes $S(b_1, \dots, b_n)$ and $R(a_1, \dots, a_m)$ if there is a grounding (of the non-free variables in) $F_{R(a_1, \dots, a_m)}$ in which the atom $S(b_1, \dots, b_n)$ occurs.

  We define the conditional probability of $R(a_1, \dots, a_m)$ given the truth values of its parent atoms to be the probability value of ${F^P}_{R(a_1, \dots, a_m)}$, which is itself defined by induction on $F_{R(a_1, \dots, a_m)}$ as follows:
    \begin{enumerate}
  \item If $F_{R(a_1, \dots, a_m)} = q$ for a $q \in [0,1]$, ${F^P}_{R(a_1, \dots, a_m)}= q$.
  \item If $F_{R(a_1, \dots, a_m)} = S(a_1, \dots, a_m)$, then ${F^P}_{R(a_1, \dots, a_m)} = 1$ if $S(a_1, \dots, a_m)$ is true and 0 otherwise.
  \item If $F_{R(a_1, \dots, a_m)} = F_1 \cdot F_2 + (1 - F_1) \cdot F_3$, then ${F^P}_{R(a_1, \dots, a_m)} = {F_1}^P \cdot {F_2}^P + (1 - {F_1}^P) \cdot {F_3}^P$.
  \item $F_{R(a_1, \dots, a_m)} = \mathrm{comb}(F_1, \dots, F_k \mid \vec{w})$, then ${F^P}_{R(a_1, \dots, a_m)} = \mathrm{comb}\{F^P\}$, where $F$ ranges over the groundings of (the variables in $\vec{w}$ in) $F_1, \dots, F_k$.
    \end{enumerate}
\end{definition}

\begin{example}
  Consider the two RBN from Example \ref{Ex_RBN}.
  In both RBN, for all elements $a$ of a given domain $D$, the events $\{S(a)\mid a \in D\}$ are independent events of probability 0.5.
  In both cases, the events $\{R(a)\mid a \in D\}$ are independent when conditioned on the set of events $\{S(a)\mid a \in D\}$.
  In $B_1$, the conditional  probability of $R(a)$ depends solely on whether $S(a)$ holds for that particular domain element (it is 0.7 if $S(x)$ holds, and 0.2 otherwise)
  and in particular the events $\{R(a)\mid a \in D\}$ are even unconditionally independent.
  In $B_2$, the conditional probability is equal to the overall proportion of domain elements $b$ for which $S(b)$ holds (the arithmetic mean of the indicator functions).
  Here, the events $\{R(a)\mid a \in D\}$ are not unconditionally independent.
\end{example}
If we are given the values of some predicates as data, these can be included as \emph{unlabelled sources}, that is, predicates with no incoming arrows and no probability functions assigned to them. In this way, RBNs also provide a way to define probability distributions over structures in a larger vocabulary given structures in a subvocabulary as data.

For instance, if the probability formula  $F_S = 0.5$ is removed from the RBNs of Example \ref{Ex_RBN}, the resulting RBNs take $\{S\}$-structures as input and return a probability distribution over expansions to $L$.

We turn to MLN:\@
\begin{definition}\label{def:MLNSyn}
  Let $\mathcal{L}$ be a vocabulary. A \emph{Markov Logic Network T }over $\mathcal{L}$
is given by a collection of pairs $\varphi_{i}:w_{i}$ (called \emph{weighted
formulas}), where $\varphi$ is a quantifier-free $\mathcal{L}$-formula
and $w\in\mathbb{R}$. We call $w$ the \emph{weight} of $\varphi$
in $T$. 
\end{definition}
\begin{example}
Consider a vocabulary with two unary relation symbols $Q$ and $R$ and
the MLN consisting of just one formula, $R(x)\wedge Q(y):w$.
Note that this is different to 
the MLN $\{R(x)\wedge Q(x):w\}$, where the variables are the same. 
\end{example}

\begin{definition}\label{def:MLNSem}
  Given a domain $D$, an MLN $T$ over $\mathcal{L}$ defines a
distribution over $D$ as follows: let $\mathfrak{X}$ be an $\mathcal{L}$-structure
on $D$. Then 
\[
\mathcal{P}_{T,D}(\mathfrak{X})=\frac{1}{Z}\exp(\underset{i}{\sum}w_{i}n_{i}(\mathfrak{X}))
\]
where $i$ varies over all the weighted formulas in $T$, $n_{i}(\mathfrak{X})$
is the number of true groundings of $\varphi_{i}$ in $\mathfrak{X}$,
$w_{i}$ is the weight of $\varphi_{i}$ and $Z$ is a normalisation
constant to ensure that all probabilities sum to 1. 
\end{definition}
\begin{example}
In the MLN $T_1 := \{R(x)\wedge Q(x):w\}$, the probability of any possible structure
$\mathfrak{X}$ with domain $D$ is proportional to $\exp\left(w\cdot n(\mathfrak{X})\right)$,
where $n(\mathfrak{X})$ is the number $\left\lvert R(x)\wedge Q(x)\right\rvert$
of elements $a$ of $D$ for which $R(a)$ and $Q(a)$ hold in the
interpretation from $\mathfrak{X}$.

In the MLN $T_2 := \{R(x)\wedge Q(y):w\}$,
however, this probability is proportional to $\exp\left(w\cdot n'(\mathfrak{X})\right)$,
where $n'(\mathfrak{X})$ is the number of pairs $(a,b)$ from $D\times D$
for which $R(a)$ and $Q(b)$ hold in the interpretation from $\mathfrak{X}$.
In other words, $n'(\mathfrak{X})$ is the product $\left\lvert R(x)\right\rvert \cdot\left\lvert Q(y)\right\rvert$.
\end{example}

The name ``Markov logic network'' is motivated by the observation that grounding to a given domain induces a Markov network in which the atoms are nodes and the edges are given by co-occurrence of two atoms in a formula.
In particular, the marginal probability of a query depends only on the connected components of the atoms occurring in that query.

Finally, we introduce probabilistic logic programs. 
\begin{definition}
  A probabilistic logic program $\Pi$ consists of a finite set of \emph{probabilistic facts}, which are expressions of the form $\alpha :: H$ for an $\alpha \in [0,1]$ and an atom $H$, and a finite set of \emph{clauses}, which are expressions of the form $H \textrm{:-} B_1, \dots, B_n$ for an atom $H$ and literals $B_1, \dots, B_n$, such that $\Pi$ is \emph{stratified}, that is, that in the directed dependency graph that has a node for every relation symbol and an edge from $S$ to $R$ if $S$ occurs in the body of a clause whose head has $R$ as its relation symbol, every edge involved in a cycle is induced by a positive occurence of $S$ (i.\ e.\ $S$ only occurs unnegated in the clause inducing the edge). We assume that there is exactly one probabilistic fact for every relation symbol that does not occur in the head of a clause. 
\end{definition}

\begin{example}\label{Ex_PLP}
  Consider the vocabulary with a binary relation symbols $R$ and $U$ and unary relation symbol $S$.
  Then one can construct the program $\Pi_1$, defined by $0.5::U(x,y)$, $0.5::S(x)$ and $R(x,y) \textrm{:-} S(x),S(y),U(x,y)$.
  Also consider the program $\Pi_2$, defined by $0.5::U(x,y)$, $0.5::R(x,y)$ and $S(x) \textrm{:-} R(x,y),U(x,y)$.  
\end{example}

The semantics of probabilistic logic programs is defined in two stages.
First, the probabilistic facts induce a distribution with respect to the subvocabulary $L'$ of those relation symbols which do not occur in the head of a clause.

\begin{definition}
  Let $\Phi$ be a finite set of probabilistic facts, whose atoms have predicates in a vocabulary $L'$.
  Let $D$ be a set.
  Then $\Phi$ defines an $L'$-distribution over $D$ given by
  independently throwing a biased coin with probability $\alpha$ and every grounding $R(\vec{a})$ of the atom of a probabilistic fact $\alpha :: R(\vec{t})$.

  In other words, only structures in which all ground atoms that are not groundings of the atom of any probabilistic fact are false are possible,
  and the probability of any possible  structure $\mathfrak{X}$ is given by
  \[
  \prod_{\substack{(\alpha :: R(\vec{t})) \in \Phi \\ R(\vec{a})\text{ grounding of }R(\vec{t})}}\alpha^{\delta(R(\vec{a}))}{\left(1-\alpha\right)}^{1 - \delta(R(\vec{a}))}
  \]
  where $\delta(R(\vec{a}))$ is 1 if $\mathfrak{X}\models R(\vec{a})$ and 0 otherwise. 
\end{definition}

The clauses now serve as a Datalog program, associating with each $L'$-structure an expansion to the full vocabulary $L$, namely their minimum Herbrand model. 
\begin{definition}
  Let $L$ be the vocabulary of all predicates occurring in a clause or probabilistic fact of a probabilistic logic program $\Pi$, and
  let $L'$ be the subvocabulary of all those predicates occurring in the atoms of probabilistic facts.

  Let $\Xi$ be the set of clauses of $\Pi$, and $\Phi$ the set of its probabilistic facts.
  Consider any  $(H \textrm{:-} B_1, \dots, B_n) \in \Xi$ as an implication $B_1 \land \dots \land B_n \rightarrow H$.
  Consider a partial order $<$ on $L$-structures $\mathfrak{Y}$ with a given domain $D$,
  where $\mathfrak{Y}_1 < \mathfrak{Y}_2$ whenever any ground atom satisfied by  $\mathfrak{Y}_1$ is also satisfied by  $\mathfrak{Y}_2$. 
  Then, since $\Pi$ is stratified,
  any $L'$-world  $\mathfrak{X}$ has a smallest expansion $\Xi(\mathfrak{X})$ to $L$ in which all the implications encoded by $\Xi$ hold \citep[Theorem 11.2]{CeriGT90}.

  Thus $\Pi$ defines an $L$-distribution over any domain $D$ by setting the probability of an $L$-structure $\mathfrak{X}$  with domain $D$ to be 0 if it is not equal to $\Xi(\mathfrak{X}_{L'})$ and
  to be the probability of $\mathfrak{X}_{L'}$ under the distribution induced by $\Phi$ otherwise. 
\end{definition}

\begin{example}
  Consider the two PLP from Example \ref{Ex_PLP} and fix a domain $D$.
  Then in $\Pi_1$, $L' = \{U,S\}$ and the induced distribution on $L'$-structures is uniform. Then the distribution is extended to $L$ through $R(x,y) \leftrightarrow S(x)\wedge S(y)\wedge U(x,y)$.
  In $\Pi_2$, $L' = \{U,R\}$ and the induced distribution on $L'$-structures is again uniform. The distribution is now extended to $L$ through $S(x) \leftrightarrow \exists_y (R(x,y) \wedge U(x,y))$.
\end{example}

Often PLP are written not only with probabilistic facts and logical clauses, but with \emph{probabilistic clauses} $C$ of the form $\alpha::H \textrm{:-} B_1, \dots, B_n$.
These are used as syntactic sugar: Let $x_1, \dots, x_n$ be the variables occurring in $H,B_1, \dots, B_n$. Then  $C$ stands for the combination of a new probabilistic fact $U_C(x_1, \dots, x_n)::\alpha$ and a clause $H \textrm{:-} B_1, \dots, B_n,U_C$.
Using this convention, one could write $\Pi_1$ with the probabilistic fact  $0.5::S(x)$ and the probabilistic clause $0.5::R(x,y) \textrm{:-} S(x),S(y)$, and
$\Pi_2$ with the probabilistic fact  $0.5::R(x,y)$ and the probabilistic clause $0.5::S(x) \textrm{:-} R(x,y)$. 

\section{Projectivity on unstructured domains}

We introduce the notion of a projective family of distributions along the lines of \cite{JaegerS18,JaegerS20}. \emph{Throughout this section, we fix a relational vocabulary $L$.}

\begin{definition}\label{JSProj}
Let $(P)$ be an $\mathbb{N}$-indexed $L$ family of distributions. 

Then $(P)$ is called \emph{exchangeable} if for any $n$, $P_n(\mathfrak{X}) = P_n(\mathfrak{Y})$ whenever $\mathfrak{X}$ and $\mathfrak{Y}$ are isomorphic $L$-structures on  $\{1, \dots, n \}$. 

$(P)$ is called \emph{projective} if it is exchangeable and for any $n' < n$ and any $L$-structure $\mathfrak{X}$ on $\{1, \dots, n' \}$, 
\[ P_{n'}(\mathfrak{X}) = P_n(\mathfrak{Y} \textrm{ extends }\mathfrak{X}).\]
\end{definition}

While this definition explicitly uses the natural numbers as representatives of the domain sizes that are ordered by inclusion, this can be avoided:

\begin{definition}\label{PureProj}
Let $(P)$ be an $L$ family of distributions.
Then $(P)$ is \emph{projective (resp.\ exchangeable)}  if for any two finite sets $D'$ and $D$, any injective (resp. bijective) map $\iota: D' \hookrightarrow D$ and any $L$-structure $\mathfrak{X}$ on $D'$ the following holds:
\[ P_{D'}(\mathfrak{X}) = P_{D}(\mathfrak{Y} \textrm{ extends }\iota(\mathfrak{X})) \]
\end{definition}

These definitions are equivalent in the following sense:

\begin{proposition}\label{FirstEqu}
  For every projective (resp.\ exchangeable) $\mathbb{N}$-indexed $L$ family of distributions $(P)$, there is a unique projective (resp.\ exchangeable) $L$ family of distributions that coincides with $(P)$ on all domains of the form $\{1,\dots,n\}$.
  Conversely, the restriction of any projective (resp.\ exchangeable) $L$ family of distributions to domains of the form $\{1, \dots, n\}$ is a projective (resp.\ exchangeable) $\mathbb{N}$-indexed $L$ family of distributions.
\end{proposition}

\begin{proof}
Let $(P)$ be an exchangeable $\mathbb{N}$-indexed $L$ family of distributions, let $D =: \{a_1, \dots, a_n\}$ be a finite set.
This leads to a bijection $f:\Omega_{\{1, \dots, n\}}^L \rightarrow \Omega_D^L$ which replaces any $i$ with $a_i$. 
Let $P_D$ be the probability distribution obtained from $P_D(\mathfrak{X}) := P_n(f^{-1}(\mathfrak{X}))$. 
Note that $f^{-1}(\mathfrak{X})$ and $\mathfrak{X}$ are isomorphic, and that therefore in particular $P_D$ is independent of the specific enumeration of $D$ by the exchangeability of the $\mathbb{N}$-indexed family $(P)$.

We show that ${(P_D)}_{D\textrm{ a finite set}}$ is exchangeable and that if $(P)$ is projective, so is ${(P_D)}_{D\textrm{ a finite set}}$. 
So let $D' =: \{b_1, \dots, b_n\}$, let $\iota:D' \rightarrow D$ be a bijective map between finite sets and let $\mathfrak{X}$ be an $L$-structure on $D'$. 
Enumerate $D =: \{a_1, \dots, a_n\}$ such that $\iota(b_i) = a_i$ for all $1\leq i \leq n$.
Let $f':\Omega_{\{1, \dots, n\}}^L \rightarrow \Omega_{D'}^L$ and $f:\Omega_{\{1, \dots, n\}}^L \rightarrow \Omega_D^L$ be the bijections induced by those enumerations. 
Then $f'^{-1}(\mathfrak{X}) = f^{-1}(\iota (\mathfrak{X}))$ and therefore $P_{D'}(\mathfrak{X}) = P_D(\iota (\mathfrak{X}))$  as required.
So assume now that $(P)$ is projective and let $\iota:D' \hookrightarrow D$ be injective. As before, we enumerate  $D' =: \{b_1, \dots, b_m\}$ and $D =: \{a_1, \dots, a_n\}$ such that $\iota(b_i) = a_i$ for all $1\leq i \leq m$. Define $f'$ and $f$ as above. 
Then $P_{D'}(\mathfrak{X}) = P_m(f'^{-1}(\mathfrak{X}))$. 
By construction,  $\{\mathfrak{Y} \in \Omega_D^L\mid\mathfrak{Y} \textrm{ extends }\iota(\mathfrak{X})\}$ are exactly those possible worlds for which $f^{-1}(\mathfrak{Y}) \in \Omega_{\{1, \dots, n\}}^L$ extends $f'^{-1}(\mathfrak{X})$. 
Therefore, the claim follows from the projectivity of $(P)$.

It remains to demonstrate the uniqueness of the extension. So let $D$ be a finite set, $\mathfrak{X}$ a possible world on $D$ and $(P_D)_{D\textrm{ a finite set}}$ an exchangeable family of distributions extending $(P_n)_{n\in \mathbb{N}}$.
Let $k$ be the cardinality of $D$. Then there is a bijection $\iota:D \rightarrow \{1, \dots, k\}$ which maps $\mathfrak{X}$ to a possible world $\iota(\mathfrak{X})$ on $\{1, \dots, k\}$.
By exchangeability, $P_D(\mathfrak{X}) = P_n(\iota(\mathfrak{X}))$, which is uniquely determined by $(P_n)_{n\in \mathbb{N}}$. 
\end{proof}

Jaeger and Schulte \cite[Section 4]{JaegerS18} identified projective fragments of RBN, MLN  and PLP (see Subsection \ref{StarAI}).

\begin{proposition}\label{fragments}
An RBN  induces a projective family of distributions if it does not contain any combination functions.

An MLN induces a projective family of distributions if it is \emph{$\sigma$-determinate} \citep{SinglaD07}, that is, if any two atoms appearing in a formula contain exactly the same variables.

A PLP induces a projective family of distributions if it is \emph{determinate} \citep{MuggletonF90,Weitkaemper21}, that is, if any variable occurring in the body of a clause also occurs in the head of the same clause. 
\end{proposition}

For the case of probabilistic logic programming, the converse holds  \cite[Theorem 31]{Weitkaemper21}:
\begin{proposition}\label{TPLP}
Every projective PLP (without function symbols, unstratified negation or higher-order constructs) is equivalent to a determinate PLP.\@
\end{proposition}

There is also a natural alternative characterisation of projectivity in terms of queries:

\begin{proposition}\label{AltClassProj}
An $L$ family of distributions is projective if and only if for every quantifier-free $L$-query $\varphi(a_1, \dots,a_m)$, the marginal probability of $\varphi(a_1, \dots,a_m)$ depends only on the $=$-type of $a_1, \dots,a_m$.
\end{proposition}

\begin{proof}
Let $(P)$ be a projective $L$ family of distributions. Then for any finite set $D$ containing $b_1, \dots, b_m$ with the same $=$-type as $a_1, \dots, a_m$, consider the injective map $\iota$ of  $a_1, \dots a_m$ into $D$ mapping $a_1, \dots, a_m$ to $b_1, \dots, b_m$ respectively. Then the $P_{\{a_1, \dots a_m\}}$-probability of $\varphi(a_1, \dots a_m)$ coincides with the $P_D$-probability of $\varphi(b_1, \dots b_m)$ by projectivity.

Conversely, let $(P)$ be a family of distributions with the property mentioned in the proposition.
Then let $D' \hookrightarrow D$ be an injective map between finite sets, $D' = \{a_1, \dots, a_m\}$ and let $\mathfrak{X}$ be an $L$-structure with domain $D'$. Let $\varphi(a_1, \dots, a_n)$ be the quantifier-free formula expressing the $L$-type of $\mathfrak{X}$. Then $P_{D'}(\mathfrak{X}) = P_{D'}(\varphi(a_1, \dots, a_n)) = P_{D}(\varphi(a_1, \dots, a_n)) = P_{D}(\mathfrak{Y}\textrm{ extends }\mathfrak{X})$.
\end{proof}

\begin{example} \label{PLPSBM}
The relational stochastic block model of Example \ref{SBM} can be expressed by the determinate ProbLog program
\begin{verbatim}
    p :: c_1(X).
    c_0(X) :- \+c_1(X).
    p_00 :: edge(X,Y) :- c_0(X), c_0(Y), X != Y.
    p_01 :: edge(X,Y) :- c_0(X), c_1(Y), X != Y.
    p_10 :: edge(X,Y) :- c_1(X), c_0(Y), X != Y.
    p_11 :: edge(X,Y) :- c_1(X), c_1(Y), X != Y.
\end{verbatim}

It therefore encodes a projective family of distributions.
Consider a quantifier-free query $\varphi(a_1,\dots,a_n)$.
To calculate the marginal probability of $\varphi(a_1,\dots,a_n)$, one first considers the
probabilities of the possible 1-types of $a_1,\dots,a_n$ consistent with $\varphi$.
Then, for any such collection of 1-types, one can calculate the conditional probability of an edge configuration consistent with $\varphi$.
Since $\varphi$ is quantifier-free, colouring and edge relation together determine whether $\varphi$ holds.
Thus, summing over the products of probability of 1-types and conditional probability of edge configuration results in the marginal probability of $\varphi$, which did not depend in any way on other information than the $a_1,\dots,a_n$ themselves, as implied by the alternative characterisation of Proposition \ref{AltClassProj}.
\end{example}

\section{Projectivity on structured domains}\label{Structured}

The concepts introduced in the preceding section are only applicable for typical statistical relational frameworks when ``the model specification does not make use of any constants referring to specific domain elements, and is not conditioned on a pre-defined structure on the domain" \citep[Section 2]{JaegerS18}. 

In this section, we overcome these limitations by allowing $L_{\mathrm{Ext}}$-structures rather than merely plain domains as input. 
This clearly suffices to allow for model specifications conditioned on a pre-defined $L_{\mathrm{Ext}}$-structure. 
In order to allow for models with named domain elements, we also allow constants in $L_{\mathrm{Ext}}$.
However, we still do not allow new constant symbols in $L_{\mathrm{Int}}$, so while the model specification might refer to given domain elements, it does not give meaning to new uninterpreted constants.

\emph{In the remainder of this paper, unless explicitly mentioned otherwise, assume that $L_{\mathrm{Ext}}$ is a (not necessarily relational) vocabulary and $L_{\mathrm{Int}} \supseteq L_{\mathrm{Ext}}$ a vocabulary extending $L_{\mathrm{Ext}}$ by additional relation symbols (but not additional constants).} 

Another very common feature of such frameworks are multi-sorted domains. For instance, a model of a university domain might distinguish between courses and persons. 
The methods of this section allow for such domains, since they can be modelled by unary $L_{\mathrm{Ext}}$ predicates.

We first introduce the basic terminology.

\begin{definition}
An $L_{\mathrm{Ext}}$-$L_{\mathrm{Int}}$ family of distributions is a map from the class of finite $L_{\mathrm{Ext}}$-structures to the class  of probability spaces, mapping a finite $L_{\mathrm{Ext}}$-structure $\mathfrak{D}$ to a probability distribution on the space $\Omega^{\mathfrak{D}}_{L_{\mathrm{Int}}}$  of $L_{\mathrm{Int}}$-structures extending $\mathfrak{D}$.
\end{definition}

On unstructured domains, an injective map conserves all the information about a tuple of elements, namely their $=$-type. On a domain which is itself an $L$-structure, the corresponding notion conserving the $L$-type of any tuple of elements is that of an embedding of $L$-structures:

\begin{definition} \label{StructProj}
An $L_{\mathrm{Ext}}$-$L_{\mathrm{Int}}$ family of distributions $(P)$ is \emph{projective} (resp. \emph{exchangeable}) if for any embedding (resp. isomorphism) $\iota : \mathfrak{D}' \hookrightarrow \mathfrak{D}$ between $L_{\mathrm{Ext}}$-structures, the following holds for all $L_{\mathrm{Int}}$-structures $\mathfrak{X}$ extending $\mathfrak{D}'$:
\[ P_{\mathfrak{D}'}(\mathfrak{X}) = P_{\mathfrak{D}}(\mathfrak{Y} \textrm{ extends }\iota(\mathfrak{X})). \]
\end{definition}

The projective fragments captured by Proposition \ref{fragments} extend to the structured case in a natural way.

\begin{proposition}\label{fragmentsext}
An RBN with vocabulary $L_{\mathrm{Int}}$ and unlabelled sources in $L_{\mathrm{Ext}}$ induces a projective $L_{\mathrm{Ext}}$-$L_{\mathrm{Int}}$ family of distributions if it does not contain any combination functions.  

A PLP with extensional vocabulary $L_{\mathrm{Ext}}$ and intensional vocabulary $L_\mathrm{Int}$ induces a projective $L_{\mathrm{Ext}}$-$L_{\mathrm{Int}}$ family of distributions if it is determinate.

A $\sigma$-determinate MLN with predicates in $L_{\mathrm{Int}}$ induces a projective  $L_{\mathrm{Ext}}$-$L_{\mathrm{Int}}$ family of distributions for any subvocabulary $L_{\mathrm{Ext}}$ of $L_{\mathrm{Int}}$. 
\end{proposition}

\begin{proof}
The proof sketches from \cite[Propositions 4.1 to 4.3]{JaegerS18} transfer verbatim to this setting.
\end{proof}

We can give a more intuitive equivalent formulation of projectivity, generalising Proposition \ref{AltClassProj}:

\begin{proposition} \label{AltNewProj}
An $L_{\mathrm{Ext}}$-$L_{\mathrm{Int}}$ family of distributions is projective if and only if for every quantifier-free $L_{\mathrm{Int}}$-query $\varphi(a_1, \dots,a_m)$, the marginal probability of $\varphi(a_1, \dots,a_m)$ depends only on the $L_{\mathrm{Ext}}$-type of $a_1, \dots,a_m$.
\end{proposition}

\begin{proof}
  Let $(P)$ be a projective $L_{\mathrm{Ext}}$-$L_{\mathrm{Int}}$ family of distributions.
  Consider the $L_{\mathrm{Ext}}$-structure $\bar{\mathfrak{D}}$ with domain $\{a_1, \dots a_m\}$, given by the type of $a_1,\dots,a_m$.
  Then for any $L_{\mathrm{Ext}}$ structure $\mathfrak{D}$ containing $b_1, \dots, b_m$ with the same $L_{Ext}$-type as $a_1, \dots, a_m$, consider the embedding $\iota$ of $\bar{\mathfrak{D}}$ into $\mathfrak{D}$ mapping $a_1, \dots, a_m$ to $b_1, \dots, b_m$ respectively.
  Then the $P_{\bar{\mathfrak{D}}}$-probability of $\varphi(a_1, \dots a_m)$ coincides with the $P_{\mathfrak{D}}$-probability of $\varphi(b_1, \dots b_m)$ by projectivity.

Conversely, let $(P)$ be a family of distributions with the property mentioned in the proposition.
Then let $\mathfrak{D}' \hookrightarrow \mathfrak{D}$ be an embedding of $L_{Ext}$-structures, $\mathfrak{D}' = \{a_1, \dots, a_n\}$ and let $\mathfrak{X}$ be an $L_{\mathrm{Int}}$ extension of $\mathfrak{D}'$. Let $\varphi(a_1, \dots, a_n)$ be the quantifier-free formula expressing the $L_{\mathrm{Int}}$-type of $\mathfrak{X}$. Then $P_{\mathfrak{D}'}(\mathfrak{X}) = P_{\mathfrak{D}'}(\varphi(a_1, \dots, a_n)) = P_{\mathfrak{D}}(\varphi(a_1, \dots, a_n)) = P_{\mathfrak{D}}(\mathfrak{Y}\textrm{ extends }\mathfrak{X})$.
\end{proof}

\begin{example} \label{StructSBM}
  The relational stochastic block model of Example \ref{PLPSBM} can be used with membership in $c_1$ as extensional data.
  It can then be expressed by the following abridged PLP.
\begin{verbatim}
    c_0(X) :- \+c_1(X).
    p_00 :: edge(X,Y) :- c_0(X), c_0(Y), X != Y.
    p_01 :: edge(X,Y) :- c_0(X), c_1(Y), X != Y.
    p_10 :: edge(X,Y) :- c_1(X), c_0(Y), X != Y.
    p_11 :: edge(X,Y) :- c_1(X), c_1(Y), X != Y.
\end{verbatim}

It therefore encodes a projective $\{c_1\}-\{c_0,c_1,edge\}$ family of distributions.

As the probablity of any edge configuration depends solely on the community membership of the nodes involved, encoded in their $\{c_0,c_1\}$-type,
the marginal probability of any quantifier-free $\{c_0,c_1,edge\}$ query can be determined from the $\{c_1\}$-type alone, corresponding to the statement of Proposition \ref{AltNewProj}
\end{example}

Proposition \ref{AltNewProj} shows that classical projectivity coincides with the new notion when $L_\mathrm{Ext} = \emptyset$.

\begin{corollary}\label{ExtCor}
An $L$ family of distributions is projective in the sense of Definition \ref{PureProj} if and only if it is projective as an $\emptyset$-$L_{\mathrm{Int}}$ family of distributions in the sense of Definition \ref{StructProj}. 
\end{corollary}

\begin{proof}
The characterisation of Proposition \ref{AltNewProj} reduces to that of Proposition \ref{AltClassProj} when $L_\mathrm{Ext} = \emptyset$.
\end{proof}

Projective families of distributions can also be combined whenever the extensional vocabulary of one and the intensional vocabulary of the other agree:

\begin{proposition}\label{Combo}
Let $(P)$ be a projective $L_{\mathrm{Ext}}$-$L$ family of distributions and $(Q)$ a projective $L$-$L_{\mathrm{Int}}$ family of distributions. 
Then $(Q \circ P)$ defined by 
\[ (Q \circ P)_\mathfrak{D} (\mathfrak{X}) := P_\mathfrak{D}(\mathfrak{X}_{L})*Q_{\mathfrak{X}_{L}}(\mathfrak{X})\]
is a projective $L_{\mathrm{Ext}}$-$L_{\mathrm{Int}}$ family of distributions. 
\end{proposition}

\begin{proof}
Let $\iota: \mathfrak{D}' \hookrightarrow \mathfrak{D}$ be an embedding of $L_{\mathrm{Ext}}$-structures and let $\mathfrak{X}$ be an $L_{\mathrm{Int}}$-structure expanding $\mathfrak{D}'$. We need to show that 
\[ (Q \circ P)_{\mathfrak{D}'}(\mathfrak{X}) = (Q \circ P)_{\mathfrak{D}}(\mathfrak{Y}\textrm{ extends }\iota(\mathfrak{X})).\]
The following calculation uses the definitions and the projectivity of $(P)$ and $(Q)$:
\begin{align*}
   {(Q \circ P)}_{\mathfrak{D}}(\mathfrak{Y}\textrm{ extends }\iota(\mathfrak{X})) = \\
   \underset{\mathfrak{Y}\textrm{ extends }\iota(\mathfrak{X})}{\sum} P_{\mathfrak{D}}(\mathfrak{Y}_L) * Q_{\mathfrak{Y}_L}(\mathfrak{Y}) = \\
   \underset{\mathfrak{Y}'\textrm{ extends }\iota(\mathfrak{X}_L)}{\sum}\left(\underset{\mathfrak{Y}_L = \mathfrak{Y}'}{\underset{\mathfrak{Y}\textrm{ extends }\iota(\mathfrak{X})}{\sum}} P_{\mathfrak{D}}(\mathfrak{Y}_L) * Q_{\mathfrak{Y}_L}(\mathfrak{Y})\right) = \\
   \underset{\mathfrak{Y}'\textrm{ extends }\iota(\mathfrak{X}_L)}{\sum} \left( P_{\mathfrak{D}}(\mathfrak{Y}') * \underset{\mathfrak{Y}_L = \mathfrak{Y}'}{\underset{\mathfrak{Y}\textrm{ extends }\iota(\mathfrak{X})}{\sum}} Q_{\mathfrak{Y}'}(\mathfrak{Y}) \right) = \\
   \underset{\mathfrak{Y}'\textrm{ extends }\iota(\mathfrak{X}_{L})}{\sum} \left(P_{\mathfrak{D}}(\mathfrak{Y}') * Q_{\mathfrak{Y}'}(\mathfrak{Y}\textrm{ extends }\iota(\mathfrak{X}))\right)  =  \\
   \underset{\mathfrak{Y}'\textrm{ extends }\iota(\mathfrak{X}_{L})}{\sum} \left(P_{\mathfrak{D}}(\mathfrak{Y}') * Q_{\mathfrak{X}_{L}}(\mathfrak{X})\right)  = \\
   P_{{\mathfrak{D}}}(\mathfrak{Y}'\textrm{ extends }\iota(\mathfrak{X}_{L})) * Q_{\mathfrak{X}_{L}}(\mathfrak{X}) = \\
   P_{{\mathfrak{D}}'}(\mathfrak{X}_{L})*Q_{\mathfrak{X}_{L}}(\mathfrak{X}) = \\
   {(Q \circ P)}_{{\mathfrak{D}}'}(\mathfrak{X}).
\end{align*}
This is exactly the desired equality. 
\end{proof}

If $L_{\mathrm{Ext}}$ is relational, consider the \emph{free} projective $L_{\mathrm{Ext}}$ family of distributions that allocates equal probability to every possible $L_{\mathrm{Ext}}$ structure on a given domain. Then every projective $L_{\mathrm{Ext}}$-$L_{\mathrm{Int}}$ family of distributions $(P)$ can be associated to the projective $L_{\mathrm{Int}}$ family of distributions obtained by concatenating it with the free projective $L_{\mathrm{Ext}}$ family of distributions. 
This will be referred to as the \emph{free completion} $\overline{(P)}$of $(P)$. By definition, for any $L_{\mathrm{Ext}}$-structure $E$ with domain $D$ and $L_{\mathrm{Int}}$-structure $\mathfrak{X}$ extending $E$,
\[ P_E(\mathfrak{X}) = \overline{P_D}(\mathfrak{Y} = \mathfrak{X}\mid\mathfrak{Y}\textrm{ expands }E).\]

  For instance, the free completion of the projective $\{c_1\}-\{c_0,c_1,edge\}$ family of distributions from Example \ref{StructSBM} is the relational stochastic block model of Example \ref{PLPSBM} where membership in both communities is equally likely.

We briefly note a partial converse of Proposition \ref{Combo}:
\begin{proposition}\label{ComboCon}
Let $L_{\mathrm{Ext}} \subseteq L \subseteq L_{\mathrm{Int}}$ and let $(P)$ be a projective $L_{\mathrm{Ext}}$-$L_{\mathrm{Int}}$ family of distributions. Then the restriction of $(P)$ to an $L_{\mathrm{Ext}}$-$L$ family of distributions $(P')$, defined by 
\[ P'_{\mathfrak{D}}(\mathfrak{X}) := P_{\mathfrak{D}}(\mathfrak{Y} \textrm{ extends }\mathfrak{X}), \]
is itself projective.
\end{proposition}

\begin{proof}
Every quantifier-free $L$-query  $\varphi$ is also an $L_{\mathrm{Int}}$ query, and the probabilities evaluated in the $L_{\mathrm{Ext}}$-$L_{\mathrm{Int}}$ and $L_{\mathrm{Ext}}$-$L$ family of distribution coincide. Then the statement follows from Proposition \ref{AltNewProj}.  
\end{proof}

On the other hand, it is generally not the case that  for any $L_{\mathrm{Ext}} \subseteq L \subseteq L_{\mathrm{Int}}$, the corresponding restriction to an $L$-$L_{\mathrm{Int}}$ family of distributions is projective too. We will investigate this in more detail in Section \ref{SigmaProj} below.

The main motivation for studying projective families of distributions lies in their excellent scaling properties, allow for marginal inference in constant time with respect to domain size \citep{JaegerS18}.

Those properties generalise directly to the new setting:

\begin{proposition}\label{StructInference}
Let $(P)$ be a projective $L_{\mathrm{Ext}}$-$L_{\mathrm{Int}}$ family of distributions. Then marginal inference with respect to quantifier-free queries (potentially with quantifier-free formulas as evidence) can be computed in constant time with respect to domain size.
\end{proposition}
\begin{proof}
  This follows immediately from Proposition \ref{AltNewProj}, as the computation can always be performed in the substructure generated by the elements mentioned in the query and the evidence. 
\end{proof}

When $L_{\mathrm{Ext}}$ is relational, Proposition \ref{Combo} and the free completion also allow the generalisation of the pertinent results from \cite{JaegerS20,Weitkaemper21} to $L_{\mathrm{Ext}}$-structured input.

\begin{proposition}\label{onlydet}
Let $L_{\mathrm{Ext}}$ be relational and let $\Pi$ be a PLP inducing a projective $L_{\mathrm{Ext}}$-$L_{\mathrm{Int}}$ family of distributions. Then $\Pi$ is equivalent to a determinate PLP. 
\end{proposition}
\begin{proof}
Consider the PLP $\Pi'$ obtained from $\Pi$ by adding the clause $\mathtt{0.5 :: R(X_1, ..., X_n)}$ for every $n$-ary extensional predicate $R$ of $\Pi$. Then $\Pi'$ induces an $\emptyset$-$L_{\mathrm{Int}}$ family of distributions given by the concatenation of $\Pi$ with the $\emptyset$-$L_{\mathrm{Ext}}$ family of distributions induced by the added clauses in isolation. 
By Proposition \ref{Combo}, $\Pi'$ induces a projective family of distributions, and by Theorem 31 of \cite{Weitkaemper21} $\Pi'$ is equivalent to a determinate PLP $\Pi_{\mathrm{d}}'$. Moreover, the probabilistic facts in $\Pi$ and $\Pi_{\mathrm{d}}'$ coincide, and the PLP $\Pi_{\mathrm{d}}$ obtained from $\Pi_{\mathrm{d}}'$ by removing the probabilistic facts introduced above is determinate and equivalent to $\Pi$.
\end{proof}

Now we consider the AHK representation of general projective families of distributions.
We augment the definition of an AHK representation \citep[Definition 6.1]{JaegerS20} to include the extensional data as part of the input.

\begin{definition}
Let $L_{\mathrm{Int}}$ be a relational vocabulary with relations of maximal arity $a \geq 1$, and let $L_{\mathrm{Ext}}$ be a subvocabulary of $L_{\mathrm{Int}}$. 

For an $n\in \mathbb{N}$, define 
$K_n := [0,1] \times \mathcal{T}_n^{L_{\mathrm{Ext}}}$.
Then an \emph{AHK model for $L_{\mathrm{Int}}$ over $L_{\mathrm{Ext}}$} is given by
\begin{enumerate}
    \item A family of i.i.d. random variables 
    \[\{U_{(i_1, \dots, i_m)} \mid i_j \in \mathbb{N}, i_1 < ... < i_m, 0 \leq m  \leq a \}\]
    uniformly distributed on $[0,1]$.
    \item A family of measurable functions 
    \[\left\{f_m : \underset{0 \leq n \leq m}{\prod} K_{n}^{\binom{m}{n}} \rightarrow \mathcal{T}_m^{L_{\mathrm{Int}} \setminus L_{\mathrm{Ext}}} {\mid} 1 \leq m \leq a \right\}.\]
    For any such $m$ and extensional $m$-type $\varphi$, we set $F_{(j_1, \dots, j_m)}(\varphi(x_1, \dots, x_m))$ to refer to the expression 
    \[ f_m\left(\Big( \big(U_{(i_1, \dots, i_n)},\varphi_{(x_{i_1}, \dots, x_{i_n})}^n\big) \Big)\right)  \]
     where the tuples to which $f$ is applied range over all strictly ascending subsequences $(i_1, \dots, i_n)$  of $(j_1, \dots, j_m)$, and are arranged in lexicographic order.
     
     We require that every $f_m$ is \emph{permutation equivariant} in the following sense:\\
     Let $\psi(x_1, \dots, x_m) := F_{(1, \dots, m)}(\varphi(x_1, \dots, x_m))$. Then for any permutation $\iota$ of $1, \dots, m$ and any extensional $m$-type $\varphi(x_1, \dots, x_m)$, 
     
    \begin{align*}
          f_m\left(\Big( \big(U_{(\iota(i_1), \dots, \iota(i_n))},\varphi_{(x_{i_1}, \dots, x_{i_n})}^n\big) \Big)\right)=\psi(x_{\iota(1)}, \dots, x_{\iota(n)})
      \end{align*} 
     where  the tuples $(i_1, \dots, i_n)$ range over all strictly ascending subsequences   of $(1, \dots, m)$, and are arranged in lexicographic order.

\end{enumerate}
\end{definition}

An AHK model over $\emptyset$ is just a reformulation of the notion of an AHK model from \cite{JaegerS20}, and we will call it an AHK model for $L$.

An AHK model represents a projective family of distributions as follows:

\begin{definition}\label{AHKind}
Let $(f_m), (U_{\vec{i}})$ be an AHK model for $L_{\mathrm{Int}}$ over $L_{\mathrm{Ext}}$. Then the distribution which assigns to every $L_{\mathrm{Ext}}$-structure ${\mathfrak{D}}$ with domain $(a_1, \dots, a_n)$ and every $L_{\mathrm{Int}}$-structure $\mathfrak{X}$ extending ${\mathfrak{D}}$ the probability of the event
\begin{equation*}
 \underset{\mathfrak{X} \models \varphi_m(a_{i_1}, \dots, a_{i_m})}{\bigwedge}\{F_{(i_1, \dots, i_m)}(\psi) = \varphi_m(x_1, \dots, x_m))\}
 \end{equation*}
 where $m$ ranges from 1 to the maximal arity of purely intensional predicates, $\varphi_m$ ranges over purely intensional data formulas of arity $m$, $(i_1, \dots, i_m)$ ranges over ascending subsequences of $(1, \dots, n)$, $\psi$ is the extensional type of $(a_{i_1}, \dots, a_{i_m})$,
 is the \emph{family of distributions induced by the AHK model}. 
\end{definition}

\begin{theorem}\label{AHK} 

Every projective $L_{\mathrm{Ext}}$-$L_{\mathrm{Int}}$ family of distributions has an AHK representation. Conversely, every family of distributions induced by an AHK representation is projective.  
\end{theorem}

\begin{proof}
  It is easy to see that every AHK representation induces a projective $L_{\mathrm{Ext}}$-$L_{\mathrm{Int}}$ family of distributions, since the probability of any quantifier-free query $\varphi$ can be computed directly from the permutation-invariant AHK functions, without regard to the remainder of the domain. 

We will now demonstrate the converse.

Consider the free completion $\overline{(P)}$, which is a projective $L_{\mathrm{Int}}$ family of distributions. 
By the main result of \cite{JaegerS20}, $\overline{(P)}$ has an AHK representation. 
We can assume that the preimage of any $L_{\mathrm{Ext}}$ datum of arity $m$ is given by an interval in $U_{i_1, \dots, i_m}$ and does not depend on any other input to the function $f_m$.

Indeed, consider the function $f'_m := \pi_m \circ f$, where $\pi_m$ is the projection from $L_{\mathrm{Int}}$-types to $L_{\mathrm{Ext}}$-types. Then $f'_m$ defines an AHK-representation for the free $L_{\mathrm{Ext}}$-family of distributions, which can also be represented by functions $g_m$ as detailed in the assumption. Therefore, $g_m = f'_m  \circ h_m$ for a measurable function $h_m$ satisfying certain requirements, and we can replace $f_m$ with $f_m \circ  h_m$ to obtain an AHK representation satisfying the assumption. \citep[Theorem 7.28]{Kallenberg05}

For every $L_{\mathrm{Ext}}$ datum $T_{\mathrm{Ext},m}$ of arity $m$, let $g_{T_{\mathrm{Ext},m}}$ be a linear bijection from $[0,1]$ to the preimage interval of  $T_{\mathrm{Ext},m}$.

Then for a world $\mathfrak{X}$ which is given by the data $(T_m)$, $P_{\mathfrak{D}}(\mathfrak{X})$ is given by $\overline{P_n}(\mathfrak{Y} = \mathfrak{X}\mid\mathfrak{Y}\textrm{ expands }{\mathfrak{D}})$, which is equivalent to  \[\mathbb{P} \left( \underset{m}{\bigwedge} \left( (U_{\vec{i}})_{\vec{i}} \in f_m^{-1}(T_m) \right) \mid \underset{m}{\bigwedge} \left( (U_{\vec{i}})_{\vec{i}} \in f_m^{-1}(T_{\mathrm{Ext},m})  \right) \right)\]
which is in turn equivalent to 
\[\mathbb{P} \left( \underset{m}{\bigwedge} \left( (U_{\vec{i}})_{\vec{i}} \in (f_m \circ g_{T_{\mathrm{Ext},m}})^{-1}(T_m) \right) \right).\]
Therefore $(f_m \circ g_{T_{\mathrm{Ext},m}})$ define an AHK representation of $(P)$.
\end{proof}

\begin{example}
  We compute the AHK representation of the relational stochastic block model of Example \ref{PLPSBM}.
  $f_1$ determines community membership, and it is independent for every node. So let $f_1(a,b) =  \mathrm{c}_1(x) \land \neg  \mathrm{c}_0(x)$ whenever $b \leq p$ and $_0 \mathrm{c}(x) \land \neg  \mathrm{c}_1(x)$ otherwise.
  $f_2$ determines the edge relations. There are four possible configurations for any pair of nodes. We give the conditions for there to be an edge in both directions; the remaining cases are analogous.
  $f_2(a,b,c,d) =  \mathrm{edge}(x,y) \land \mathrm{edge}(y,x)$ if $b \leq p$,  $c \leq p$ and $d \leq {p_{11}}^2$, $b \leq p$,  $c > p$ and $d \leq p_{01}p_{10}$,  $b > p$,  $c \leq p$ and $d \leq p_{01}p_{10}$, or if  $b > p$,  $c > p$ and $d \leq {p_{00}}^2$.

  Now consider the relational stochastic block model with extensional community membership from Example \ref{StructSBM}.
  Then whether $\mathrm{c}_0$ holds depends entirely on whether $\mathrm{c}_1$ holds as part of the extensional data.
  Thus, $f_1(a,b,\varphi) =  \mathrm{c}_0$ if $\varphi$ entails $\neg \mathrm{c}_1(x)$ and $f_1(a,b,\varphi) = \neg \mathrm{c}_0$ otherwise.
  The dependence in $f_2$ on $b$ and $c$ are now replaced with direct dependence on the extensional community structure:
  $f_2(a,b,c,d,\varphi) =  \mathrm{edge}(x,y) \land \mathrm{edge}(y,x)$ if $\varphi$ entails $\mathrm{c}_1(x)$ and $\mathrm{c}_1(y)$ and $d \leq {p_{11}}^2$, $\varphi$ entails $\mathrm{c}_1(x)$ and $\neg \mathrm{c}_1(y)$ and $d \leq p_{01}p_{10}$,  $\varphi$ entails $\neg \mathrm{c}_1(x)$ and $\mathrm{c}_1(y)$ and $d \leq p_{01}p_{10}$, or if $\varphi$ entails $\neg \mathrm{c}_1(x)$ and $\neg \mathrm{c}_1(y)$ and $d \leq {p_{00}}^2$.

  Note that in both cases, there was no dependence on the random variable $U_{\emptyset}$, the first entry in the function signatures of $f_1$ and $f_2$.
  By including such a dependence, one can express finite or infinite mixtures of relational stochastic block models.
  With the same arguments as \cite{Weitkaemper21}, Proposition \ref{onlydet} here implies that such infinite mixtures are not expressible by a probabilistic logic program. 
\end{example}

The AHK representation allows us to derive an invariance property for projective $L_{\mathrm{Ext}}$-$L_{\mathrm{Int}}$ families of distributions, which limits the interaction between extensional predicates and output probabilities to the arity of the intensional predicates:

\begin{corollary}\label{aritylimit}
Let $(P)$ be a projective $L_{\mathrm{Ext}}$-$L_{\mathrm{Int}}$ family of distributions, and let $\varphi$ be a quantifier-free $L_{\mathrm{Int}}$-query with literals of arity at most $m$. Then let ${\mathfrak{D}}$ and ${\mathfrak{D}}'$ be $L_{\mathrm{Ext}}$-structures on the same domain which coincide on the interpretation of $L_{\mathrm{Ext}}$-literals of arity not exceeding $m$. Then $P_{\mathfrak{D}}(\varphi) = P_{{\mathfrak{D}}'}(\varphi)$. 
\end{corollary}

\begin{proof}
Consider the AHK representation $\vec{f}$ of $(P)$. As the truth value of $\varphi$ depends only on the data of arity less than or equal to $m$, it is determined by the values of $(f_i)_{i \in 1, \dots, m}$. However, none of these functions take extensional data of arity more than $m$ as arguments, and therefore the induced functions on the random variables $(U_{\vec{i}})_{i \in 1, \dots, m}$ coincide for ${\mathfrak{D}}$ and ${\mathfrak{D}}'$.  
\end{proof}

In light of Corollary \ref{aritylimit}, let us consider the scenarios of Examples \ref{lead}.\ref{multilink} and \ref{lead}.\ref{epidemiology} and evaluate the plausibility of projective modelling:

\begin{example}
In mining multiple networks, the extensional predicates are of arities 1 (node attributes) and 2 (node connections in other networks), while the intensional predicate is of arity 2 (node links in this network).
This could be expressed by a projective family of distributions in which the representing function $f_2$ depends on all the available extensional data.

Contrast this with the epidemiological case, where the extensional predicate is of arity 2 (social connections) while the intensional predicate is of arity 1 (illness of a node individual). In this case, Corollary \ref{aritylimit} implies that in a projective model, the inter-node connections have no impact on illness in the population. This goes against the modelling intention, so that a projective family with structured input is unlikely to be adequate for this domain. 
\end{example}

\section{Projectivity and infinite domains}\label{Infinite}

There has been significant work on statistical relational formalisms for infinite domains. In the context of RBNs, this was considered by Jaeger \cite{Jaeger98a}, and in the context of MLNs, by Singla and Domingos \cite{SinglaD07}. 

The first hurdle to considering infinite domains is that there are uncountably many possible worlds with a given infinite domain $D$ and a given vocabulary. 
Therefore, we need to take care in defining the $\sigma$-algebra of sets of structures to which we allocate a probability. 
We consider the \emph{local $\sigma$-algebra}. 
That is the $\sigma$-algebra generated by the sets $D_\mathfrak{X}$ of all possible worlds extending $\mathfrak{X}$, where $\mathfrak{X}$ ranges over all possible worlds whose domain is a finite subset of $D$.
This is equivalent to the \emph{event space} of \cite{Georgii11,SinglaD07}.    

We abuse notation by calling probability measures on this measure space \emph{$L$-distributions over $D$}. 
Such a distribution $P$ is called \emph{exchangeable} if for any permutation $\iota$ of $D$ and all  possible worlds $\mathfrak{X}$ whose domain is a finite subset of $D$ , $P(D_{\mathfrak{X}}) = P(D_{\iota(\mathfrak{X})})$. 

With these preliminaries, we obtain the following statement:

\begin{proposition}\label{ProjectivePureInf}
There is a one-to-one relationship between exchangeable $L$-distributions on $\mathbb{N}$ and projective $L$ families of distributions, induced by the equation
\[ P(\mathbb{N}_{\mathfrak{X}}) = P_D({\mathfrak{X}}) \] 
for any generator $\mathbb{N}_{\mathfrak{X}}$ of the local $\sigma$-algebra on $\mathbb{N}$, where $D$ is the domain of $\mathfrak{X}$.
\end{proposition}

\begin{proof}
This is a direct consequence of Kolmogorov's Extension Theorem \citep[Theorem 6.16]{Kallenberg02}, and can also be obtained as a special case of Theorem \ref{ProjectiveFraisse} below. 
\end{proof}

We briefly outline some implications of Proposition \ref{ProjectivePureInf} for studying infinite statistical relational models. 
Singla and Domingos \cite{SinglaD07} use Gibbs measure theory to show that $\sigma$-determinate MLNs give well-defined probability distributions on infinite domains. 
This also follows immediately from Proposition  \ref{ProjectivePureInf} and the projectivity of $\sigma$-determinate MLNs.  

More generally, Proposition \ref{ProjectivePureInf} lets us transfer the complete characterisation of projective families in terms of AHK representations to exchangeable distributions on the countably infinite domain.   

\begin{corollary}\label{AHKinf}
An $L$-distribution $P$ on $\mathbb{N}$ is exchangeable if and only if it has an AHK representation, that is, an AHK model for $L$ such that $P(\mathbb{N}_{\mathfrak{X}})$ is given by Definition \ref{AHKind}. 
\end{corollary}

We continue by investigating the relationship between infinite domains and projective $L_{\mathrm{Ext}}$-$L_{\mathrm{Int}}$ families of distributions.

In this case, there is no longer a unique type of infinite domain, since there are in fact uncountably many nonisomorphic countable $L_{\mathrm{Ext}}$-structures. However, for $L_{\mathrm{Ext}}$ without constants (or propositions), we can use the \emph{generic structure} or \emph{Fra\"{i}ss\'{e} limit} of the vocabulary.

We briefly summarise the relevant theory \citep{Hodges93}:
For every relational vocabulary $L_{\mathrm{Ext}}$ and every $L_{\mathrm{Ext}}$ sentence $\varphi$, let $p_{\varphi}(n)$ be the fraction of possible $L_{\mathrm{Ext}}$ worlds on domain $\{1, \dots, n\}$ which satisfy $\varphi$. 
Then by the well-known 0-1 theorem of finite model theory, 
\[ \underset{n \rightarrow \infty}{\mathrm{lim}} p_{\varphi}(n) \in \{0,1\}\]
for every $L_{\mathrm{Ext}}$ sentence $\varphi$. 
The first-order theory of all sentences whose probabilities limit to 1 has a unique countable model up to isomorphism, called the \emph{generic structure of $L_{\mathrm{Ext}}$}. 

This model has the following property, a characterisation known as Fra\"{i}ss\'{e}'s Theorem 

\begin{proposition}
Let $\mathfrak{U}$ be the generic structure of a relational vocabulary $L_{\mathrm{Ext}}$. 
Then every countable $L_{\mathrm{Ext}}$-structure ${\mathfrak{D}}$ can be embedded in $\mathfrak{U}$, and if ${\mathfrak{D}}$ is finite, then whenever $\iota_1$ and $\iota_2$ are two embeddings of ${\mathfrak{D}}$ into $\mathfrak{U}$, there is an automorphism $f$ of $\mathfrak{U}$ such that $f \circ \iota_1 = \iota_2$.  
\end{proposition}
\begin{proof}
  A good exposition of the whole theory of Fra\"{i}ss\'{e} limits can be found in Chapter 7.1 of  Hodges' textbook \cite{Hodges93}, where all the references in this proof refer to.
  The generic structure is derived as a Fra\"{i}ss\'{e} limit on pages 352-353.
  More particularly, the proposition at hand can be derived as follows.
  
  Consider the class of all finite  $L_{\mathrm{Ext}}$-structures. This class has a unique Fra\"{i}ss\'{e} limit, that is, a  countable $L_{\mathrm{Ext}}$-structure with the properties of the proposition.
  This follows from Theorem 7.1.2, with the statement on countable models a special case of  Lemma 7.1.3. Lemma 7.4.6 asserts that the Fra\"{i}ss\'{e} limit indeed coincides with the generic structure.
  
\end{proof}

\begin{example}
  Consider the case of directed graphs, that is, a single binary relation $E$.
  In this case, the generic model is a directed version of the \emph{Rado graph}.
  It can be obtained in various alternate ways; for instance, it is the graph
  obtained with probability 1 when throwing a fair coin for any pair of natural numbers $(m,n)$ and
  drawing an arc from $m$ to $n$ if the coin shows heads.

  It is also characterised by the \emph{extension axioms}, which say that for any finite subgraph on nodes $(a_1,\dots,a_n)$
  possible configuration of edges on nodes $(a_1,\dots,a_n,y_1,\dots,y_m)$ extending the known configuration of  $(a_1,\dots,a_n)$, there are $(b_1,\dots,b_m)$ in the Rado graph such that
  $(a_1,\dots,a_n,b_1,\dots;b_m)$ have the prescribed configuration.

  For other signatures than a single binary relation, analogous characterisations hold. 

\end{example}

We generalise our notions to this new setting of a single infinite structure as domain.

\begin{definition}
  Let $\mathfrak{D}$ be a countably infinite $L_{\mathrm{Ext}}$-structure.
  Then the \emph{local $\sigma$-algebra} on  $\mathfrak{D}$ is generated by the sets $\mathfrak{D}_\mathfrak{X}$ of all possible expansions of $\mathfrak{D}$ to  $L_{\mathrm{Int}}$ extending $\mathfrak{X}$, where $\mathfrak{X}$ ranges over all  $L_{\mathrm{Int}}$-structures expanding a finite substructure of $\mathfrak{D}$.

We then call probability measures on this measure space \emph{$L_{\mathrm{Int}}$-distributions over $\mathfrak{D}$}. 
Such a distribution $P$ is called \emph{exchangeable} if for any automorphism $\iota$ of $\mathfrak{D}$ and all  $L_{\mathrm{Int}}$-structures expanding a finite substructure of $\mathfrak{D}$,
$P(\mathfrak{D}_{\mathfrak{X}}) = P(\mathfrak{D}_{\iota(\mathfrak{X})})$. 
\end{definition}

We can now generalise Proposition \ref{ProjectivePureInf} to projective families with structured input:

\begin{theorem}\label{ProjectiveFraisse}
There is a one-to-one relationship between projective $L_{\mathrm{Ext}}$-$L_{\mathrm{Int}}$ families of distributions $(P)$ and exchangeable $L_{\mathrm{Int}}$-distributions $P$ over the Fra\"{i}ss\'{e} limit $\mathfrak{U}$ of $L_{\mathrm{Ext}}$, induced by the equation
\[P(\mathfrak{U}_{\mathfrak{X}}) = P_{\mathfrak{X}_{L_{\mathrm{Ext}}}}(\mathfrak{X}).\]
for any generator $\mathfrak{U}_{\mathfrak{X}}$ of the local $\sigma$-algebra on $\mathfrak{U}$.
\end{theorem}

To improve readability, the proof is postponed to the next subsection.

However, the properties of Fra\"{i}ss\'{e} limits allow even more -- every exchangeable family of distributions there can be extended to a projective family of distributions on all countable structures.

\begin{definition}
An \emph{$L_{\mathrm{Ext}}$-$L_{\mathrm{Int}}$ family of distributions  $(P)$ on countable structures} is a map taking countable $L_{\mathrm{Ext}}$-structures ${\mathfrak{D}}$ as input and returning distributions over ${\mathfrak{D}}$.  
$(P)$ is \emph{projective} (resp. \emph{exchangeable}) if for every embedding (resp. isomorphism) $\iota:{\mathfrak{D}}' \hookrightarrow {\mathfrak{D}}$ between countable $L_{\mathrm{Ext}}$-structures and every $L_{\mathrm{Int}}$-structure $\mathfrak{X}$ expanding a finite substructure of ${\mathfrak{D}}'$, 
\[ P_{{\mathfrak{D}}'}({\mathfrak{D}}'_{\mathfrak{X}}) = P_{{\mathfrak{D}}}({\mathfrak{D}}_{\iota(\mathfrak{X})}). \]
\end{definition}

\begin{theorem}\label{ProjectiveInfinite}
  Every exchangeable $L_{\mathrm{Int}}$-distribution over the Fra\"{i}ss\'{e} limit $\mathfrak{U}$ of $L_{\mathrm{Ext}}$ extends uniquely to a projective $L_{\mathrm{Ext}}$-$L_{\mathrm{Int}}$ family of distributions on countable structures. \end{theorem}

\begin{proof}
For any  countable $L_{\mathrm{Ext}}$-structure ${\mathfrak{D}}$ let $f_{\mathfrak{D}}:{\mathfrak{D}} \hookrightarrow \mathfrak{U}$ be an embedding into the Fra\"{i}ss\'{e} limit. 
Let $\mathfrak{X}$ be an $L_{\mathrm{Int}}$-structure expanding a finite $L_{\mathrm{Ext}}$-substructure ${\mathfrak{D}}'$ of  $\mathfrak{D}$. 
Then set $P_{\mathfrak{D}}({\mathfrak{D}}_{\mathfrak{X}}) := P_{\mathfrak{U}}(\mathfrak{U}_{f_{\mathfrak{D}}(\mathfrak{X})})$. 
This is well-defined, since $f_{\mathfrak{D}}$ restricts to an embedding from ${\mathfrak{D}}'$ into $\mathfrak{U}$ and any two embeddings from ${\mathfrak{D}}'$ to $\mathfrak{U}$ are conjugated by an automorphism of $\mathfrak{U}$. 
We need to show that $(P_{\mathfrak{D}})$ is a projective family of distributions on countable structures. So let $\iota:{\mathfrak{D}}' \hookrightarrow {\mathfrak{D}}$ be an embedding between countable $L_{\mathrm{Ext}}$-structures and let $\mathfrak{X}$ be an $L_{\mathrm{Int}}$-structure expanding a finite $L_{\mathrm{Ext}}$-substructure of ${\mathfrak{D}}'$.  Then 
\begin{align*}
  P_{\mathfrak{D}}({\mathfrak{D}}_{\iota(\mathfrak{X})}) = P_{\mathfrak{U}}(\mathfrak{U}_{f_{{\mathfrak{D}}} \circ \iota(\mathfrak{X})})=
  = P_{\mathfrak{U}}(\mathfrak{U}_{f_{{\mathfrak{D}}'}(\mathfrak{X})}) = P_{{\mathfrak{D}}'}({\mathfrak{D}}'_{\mathfrak{X}})
  \end{align*}
as required.
\end{proof}

Theorem \ref{ProjectiveInfinite} allows us to define projective families of distributions on the uncountable set of countable $L_{\mathrm{Ext}}$-structures by a single probability distribution on a single measure space. 
Together with Theorem \ref{ProjectiveFraisse} it implies that every projective family of distributions on finite structures can be uniquely extended to projective families of distributions on infinite structures.

\begin{corollary}
Every projective $L_{\mathrm{Ext}}$-$L_{\mathrm{Int}}$ family of distributions extends uniquely to a projective $L_{\mathrm{Ext}}$-$L_{\mathrm{Int}}$ family of distributions on countable structures.
\end{corollary}

We sketch an example of applying this theorem, which also serves to illustrate the importance of projectivity in the context of infinite models.
\begin{example}
  Consider unary predicate symbols $R_1,\dots,R_n,P$, let $L_{\mathrm{Ext}} := \{R_1,\dots,R_n\}$ and let $L_{\mathrm{Int}} := \{R_1,\dots,R_n,P\}$.
  Then a projective $L_{\mathrm{Ext}}$-$L_{\mathrm{Int}}$ family of distributions on countable structures
  can be used
to model a dynamic system in which some attribute $P(t)$ varies stochastically depending on observed attributes $R_1(t),\dots,R_n(t)$.

So assume that one has one or more simulations of possible developments of  $R_1(t),\dots,R_n(t)$ over time as well as possibly some data on the previous development of $P(t)$ and  $R_1(t),\dots,R_n(t)$ compatible with those models.

Then such a projective family over countable structures allows one to pose various queries of interest about $P$, from asking about certain time points
(``What is the likelihood of $P(1000)$ if  $R_1,\dots,R_n$ develop in this way?'') to asking about the long term structure of the process (``What is the likelihood that $P(t)$ will hold at infinitely many time points $t$ if  $R_1,\dots,R_n$ develop in this way?''). Of course, all such queries can be conditioned on the observed previous development, which simply means conditioning on a certain initial segment

Not only are such queries well-defined, but the projectivity of the family means that query probabilities are preserved under embeddings.
For instance, assume we increase the sampling frequency of the simulation, so instead of a domain of $t = \{10,20,30, \dots\}$, say, we transition to a domain of $t = \{1,2,3, \dots\}$.
Then projectivity ensures that when restricted to time points divisible by 10, the answers to the queries above will remain unchanged.  
\end{example}

\subsection*{Proof of Theorem \ref{ProjectiveFraisse}}

The proof of Theorem \ref{ProjectiveFraisse} rests on two technical lemmas on a generating subset of the local $\sigma$-algebra.

\begin{definition}
  Let $\mathfrak{X}$ be a countably infinite structure and fix an enumeration  $\{a_1, a_2, \dots\}$ of the elements of the domain of $\mathfrak{X}$. Then an \emph{initial segement} of  $\mathfrak{X}$ is a substructure $\mathfrak{Y}$ of $\mathfrak{X}$ whose domain  is of the form $\{a_1,\dots,a_n\}$  for an $n \in \mathbb{N}$.
\end{definition}

\begin{lemma}\label{initialsegs}
  Fix any enumeration $\{a_1, a_2, \dots\}$ of the elements of the domain $D$ of $\mathfrak{U}$. 
  Then the local $\sigma$-algebra on $\mathfrak{U}$ is generated by the subset of those $\mathfrak{U}_{\mathfrak{X}}$ for which $\mathfrak{X}_{L_{\mathrm{Ext}}}$ is an initial segment of $\mathfrak{U}$. 
\end{lemma}

\begin{proof}
  Let $\mathfrak{X}$ be an expansion to $L_{\mathrm{Int}}$ of a finite substructure of $\mathfrak{U}$, and let $a_n$ be the element of highest index in the domain of $\mathfrak{X}$. 
  Let $\{\mathfrak{X}_i\}_{i \in I}$ be the set of all extensions of $\mathfrak{X}$ to the domain $\{a_1, \dots, a_n\}$. Then $D_{\mathfrak{X}} = \underset{i \in I}{\bigcup}D_{\mathfrak{X}_i}$ as required.
\end{proof}

\begin{lemma} \label{unionlemma}
  Let ${\mathfrak{X}}$ and $\{{\mathfrak{X}_i}\}_{i \in I}$ be expansions to $L_{\mathrm{Int}}$ of initial segments of $\mathfrak{U}$ under some ordering of the domain of $\mathfrak{U}$. 
  If $\mathfrak{U}_{\mathfrak{X}}$ is the union of $\{\mathfrak{U}_{\mathfrak{X}_i}\}_{i \in I}$, then there is a finite subset $I' \subseteq I$ such that $\mathfrak{U}_{\mathfrak{X}}$ is the union of $\{\mathfrak{U}_{\mathfrak{X}_i}\}_{i \in I'}$. 
\end{lemma}

\begin{proof}
  Consider the tree $G$ whose nodes are expansions  $\mathfrak{Y}$ to  $L_{\mathrm{Int}}$ of initial segments of $\mathfrak{U}$ that extend $\mathfrak{X}$, but do not extend any $\mathfrak{X}_i$. Let there be an edge from $\mathfrak{Y}$ to $\mathfrak{Y}'$ in $G$ whenever $\mathfrak{Y}'$ extends $\mathfrak{Y}$ by a single element. 
  If $G$ is empty, $\mathfrak{X}$ itself extends an $\mathfrak{X}_i$, and we can choose $I=\{i\}$. So assume that $G$ is non-empty. Then $\mathfrak{X}$ is the root of $G$. Furthermore, every level of $G$ is finite, since there are only finitely many possible expansions of any (finite) initial segment of $\mathfrak{U}$ to $L_{\mathrm{Int}}$.

  We show that $G$ is finite. Assume not. Then by K\"onig's Lemma \citep[III.5.6]{Kunen11} we can conclude that there is an infinite branch $\rho$ in $G$. Consider the structure $\mathfrak{Z} := \bigcup \rho$. Since $\rho$ is infinite, $\mathfrak{Z}$ expands $\mathfrak{U}$. Additionally,  $\mathfrak{Z}$ does not extend any $\mathfrak{X}_i$ by construction. Therefore, $\mathfrak{Z} \in \mathfrak{X} \setminus \underset{i \in I}{\bigcup} \mathfrak{X}_i$, \emph{contradicting} the assumption of the lemma.   

  So $G$ is finite. Let $n$ be the cardinality of the largest $\mathfrak{Y} \in G$. Then choose $I'$ to be those $i \in I$ whose cardinality does not exceed $n$. By the definition of $G$,  $D_{\mathfrak{X}}$ is the union of $\{D_{\mathfrak{X}_i}\}_{i \in I'}$ as required.
\end{proof}

Now we proceed to the proof of Theorem \ref{ProjectiveFraisse}.
\begin{proof}
  Let $(P)$ be a projective $L_{\mathrm{Ext}}$-$L_{\mathrm{Int}}$ family of distributions.
  We show that $P(\mathfrak{U}_{\mathfrak{X}}) := P_{\mathfrak{X}_{L_{\mathrm{Ext}}}}(\mathfrak{X})$ defines an exchangeable family of distributions on $\mathfrak{U}$. 
  Fix an enumeration of $\mathfrak{U}$. 
  Let $\mathcal{U}$ be the class of all sets of the form  $\mathfrak{U}_{\mathfrak{X}}$, where $\mathfrak{X}$ is an expansion of an initial segment of $\mathfrak{U}$.
  We recall the definition of a semiring of sets:

  A \emph{semiring of sets} \citep[I.5.1]{Elstrodt11} is a class of sets $\mathcal{C}$ with the following properties:
  \begin{enumerate}
  \item The empty set is contained in $\mathcal{C}$. 
  \item $\mathcal{C}$ is closed under finite intersections.
  \item For any $X,Y \in \mathcal{C}$, $Y \setminus X$ is a finite union of sets in $\mathcal{C}$.
  \end{enumerate}

  We show that $\mathcal{U}$ forms a semiring of sets . 
  Indeed, the empty set lies in $\mathcal{U}$ by construction.
  Let $\mathfrak{U}_{\mathfrak{X}},\mathfrak{U}_{\mathfrak{Y}} \in \mathcal{U}$, and without loss of generality let the domain of $\mathfrak{X}$ be contained in the domain of $\mathfrak{Y}$. 
  Then if $\mathfrak{Y}$ extends $\mathfrak{X}$,  $\mathfrak{U}_{\mathfrak{Y}} \subseteq \mathfrak{U}_{\mathfrak{X}}$ and thus $\mathfrak{U}_{\mathfrak{X}} \cap \mathfrak{U}_{\mathfrak{Y}} = \mathfrak{U}_{\mathfrak{Y}}$. 
  If  $\mathfrak{Y}$ does not extend $\mathfrak{X}$, then no expansion of $\mathfrak{U}$ can simultaneously extend $\mathfrak{X}$ and $\mathfrak{Y}$, so $\mathfrak{U}_{\mathfrak{X}} \cap \mathfrak{U}_{\mathfrak{Y}} = 0$. 
  Similarly, if $\mathfrak{Y}$ does not extend $\mathfrak{X}$, then $\mathfrak{U}_{\mathfrak{Y}} \setminus \mathfrak{U}_{\mathfrak{X}} = \mathfrak{U}_{\mathfrak{Y}}$ and $\mathfrak{U}_{\mathfrak{X}} \setminus \mathfrak{U}_{\mathfrak{Y}} = \mathfrak{U}_{\mathfrak{X}}$, while if $\mathfrak{Y}$ does extend $\mathfrak{X}$, $\mathfrak{U}_{\mathfrak{Y}} \setminus \mathfrak{U}_{\mathfrak{X}} = \emptyset$. 
  So assume that $\mathfrak{Y}$ extends $\mathfrak{X}$. 
  Then the difference $\mathfrak{U}_{\mathfrak{X}} \setminus \mathfrak{U}_{\mathfrak{Y}}$ is given by the union $\bigcup \mathfrak{U}_{\mathfrak{Y}_i}$, where $\mathfrak{Y}_i$ ranges over all expansions of $\mathfrak{Y}_{L_{\mathrm{Ext}}}$ extending $\mathfrak{X}$ which are not equal to $\mathfrak{Y}$. This is a disjoint union of sets in $\mathcal{U}$ as required.  

  We show that $P$ defines a premeasure on this semiring. 
  Then by Caratheodory's Extension Theorem \citep[II.4.5]{Elstrodt11}, $P$ extends to a measure on the generated $\sigma$-algebra, which coincides with the local $\sigma$-algebra by Lemma \ref{initialsegs}. 
  $P$ is clearly semipositive, and $P(\emptyset) = 0$ and $P(\mathfrak{U}) = 1$ by construction. 
  It remains to show that $P$ is $\sigma$-additive. 
  So let $\mathfrak{U}_{\mathfrak{X}}$ be the disjoint union of $\{\mathfrak{U}_{\mathfrak{X}_i}\}_{i \in I}$. 
  By Lemma  \ref{unionlemma}, we can assume without loss of generality that $I$ is finite. 
  Let $a_n$ be the element of highest index in the domain of any of $\mathfrak{X}$ and the $\{\mathfrak{U}_{\mathfrak{X}_i}\}_{i \in I}$. 
  Let $A_n$ be the initial segment of $\mathfrak{U}$ of length $n$.
  Since $P_{A_n}$ is additive, 
  \[P_{A_n}(\mathfrak{Y}\textrm{ extends }\mathfrak{X}) = \underset{i \in I}{\Sigma}P_{A_n}(\mathfrak{Y}\textrm{ extends }\mathfrak{X}_i)\]
  and by projectivity 
  \[P_{A_n}(\mathfrak{Y}\textrm{ extends }\mathfrak{X}) = P_{\mathfrak{X}_{L_{\mathrm{Ext}}}}(\mathfrak{X})\] 
  and
  \[P_{A_n}(\mathfrak{Y}\textrm{ extends }\mathfrak{X}_i) = P_{{\mathfrak{X}_i}_{L_{\mathrm{Ext}}}}(\mathfrak{X}_i)\] 
  for every $i \in I$. 
  This shows that $P(\mathfrak{U}_{\mathfrak{X}}) := P_{\mathfrak{X}_{L_{\mathrm{Ext}}}}(\mathfrak{X})$ defines a probability distribution over $\mathfrak{U}$. 

  To show exchangeability, consider an automorphism $\iota$ of $\mathfrak{U}$ and an expansion $\mathfrak{X}$ of a finite substructure of $\mathfrak{U}$. Then 
  \[P(\mathfrak{U}_{\mathfrak{X}}) = P_{\mathfrak{X}_{L_{\mathrm{Ext}}}}(\mathfrak{X}) = P_{\iota(\mathfrak{X}_{L_{\mathrm{Ext}}})}(\iota(\mathfrak{X})) = P(\mathfrak{U}_{\iota(\mathfrak{X})})\] as required. 

  Conversely let $P$ be an exchangeable distribution over $\mathfrak{U}$. 
  $\mathfrak{U}$ is the generic structure of $L_{\mathrm{Ext}}$. 
  Thus, for any finite $L_{\mathrm{Ext}}$-structure $A$, there is an embedding $f:A \hookrightarrow \mathfrak{U}$, and if $f_1,f_2$ are two such embeddings, there is an automorphism $g$ of $\mathfrak{U}$ such that $f_2 = g \circ f_1$. 
  Define $P_A(\mathfrak{X}) := P(\mathfrak{U}_{f(\mathfrak{X})})$ for any finite $L_{\mathrm{Ext}}$-structure $A$ and any expansion $\mathfrak{X}$ of $A$ to $L_{\mathrm{Int}}$. Since $P$ is an exchangeable distribution over $\mathfrak{U}$, $P_A$ is well-defined and itself a probability distribution.
  We proceed to show that $(P_A)$ defines a projective family of distributions. 
  So let $\iota:A' \hookrightarrow A$ be an embedding of $L_{\mathrm{Ext}}$-structures. Let $f$ be an embedding of $A$ into $\mathfrak{U}$. 
  Then $f' := f \circ \iota$ is an embedding of $A'$ into $\mathfrak{U}$.
  Let $\mathfrak{X}$ be an expansion of $A'$ to $L_{\mathrm{Int}}$.
  We need to verify that 
\[P_{A'}(\mathfrak{X}) = P_A(\mathfrak{Y}\textrm{ extends }\iota(\mathfrak{X})).\]
By definition, $P_{A'}(\mathfrak{X}) = P(\mathfrak{U}_{f \circ \iota (\mathfrak{X})})$.
Also by definition, $P_A(\mathfrak{Y}\textrm{ extends }\iota(\mathfrak{X}))$ is given by \[P\left(\underset{\mathfrak{Y}\textrm{ extends }f\circ \iota(\mathfrak{X})\textrm{ to }f(A)}{\bigcup} \mathfrak{U}_{\mathfrak{Y}}\right) = P(\mathfrak{U}_{f \circ \iota (\mathfrak{X})})\]
\end{proof}

\section{$\sigma$-projectivity}\label{SigmaProj}
Even though projective families of distributions allow scaling with domains in the original mode, this is not necessarily preserved if some predicates are treated as observed:

\begin{example} \label{sigmaSBM}
  Consider the relational stochastic block model of Example \ref{SBM}.
  We saw there that it is projective when considered as  an $L$-family of distributions, where $L$ includes both the community relation and the edge relation. 
  However, when treated as an $\{\mathrm{edge}\}$-$\{\mathrm{edge},\mathrm{community}\}$ family, that is, a model for predicting community membership in which the edge relation is given as data, the model is no longer projective.
  This can be seen by assuming $p_{11}$ and $p_{10}$ to be larger than $p_{01}$ and $p_{00}$ respectively. Then, the existence of any edge away from a node increases the likelihood of that node lying in Community 1. Thus, the likelihood of a node depends not merely on the quantifier-free  $\{\mathrm{edge}\}$-type of the single node but also on its relationship to other nodes, violating projectivity. 
\end{example}

We call those families $\sigma$-projective, where projectivity is preserved under treating any subvocabulary as data. More precisely:

\begin{definition}
A projective  $L_{\mathrm{Ext}}$-$L_{\mathrm{Int}}$ family of distributions $(P)$ is called \emph{regular} if for any finite $L_{\mathrm{Int}}$-structure $\mathfrak{X}$, $P_{ \mathfrak{X} _{L_{\mathrm{Ext}}}}(\mathfrak{X})>0$.

If $L_{\mathrm{Ext}} \subseteq L \subseteq L_{\mathrm{Int}}$, a regular projective $L_{\mathrm{Ext}}$-$L_{\mathrm{Int}}$ family of distributions gives rise to an $L$-$L_{\mathrm{Int}}$ family of distributions by setting
\[P_{\mathfrak{X}_L}(\mathfrak{X}) := P_{\mathfrak{X}_{L_{\mathrm{Ext}}}}(\mathfrak{Y} = \mathfrak{X}\mid \mathfrak{Y}_L=\mathfrak{X}_L).\] 
$(P)$ is called \emph{$\sigma$-projective} if for any such $L$ the associated $L$-$L_{\mathrm{Int}}$ family of distributions is again projective. 
\end{definition}

Paradigmatic examples of $\sigma$-projective families are those induced by $\sigma$-determinate MLNs.

\begin{proposition}\label{sigmasigma}
The family of distributions induced by a $\sigma$-determinate MLN is $\sigma$-projective.
\end{proposition}

\begin{proof}
This follows immediately from Proposition \ref{fragmentsext}.
\end{proof}

Proposition \ref{sigmasigma} has implications for the expressivity of $\sigma$-determinate MLNs:
\begin{corollary}
The relational stochastic block model of Example \ref{sigmaSBM} cannot be expressed by a $\sigma$-determinate MLN.\@ 
\end{corollary}

On the other hand, determinate ProbLog programs or RBNs without combination functions are not $\sigma$-projective in general, since both formalisms can express the stochastic blocks model. 

Malhotra and Serafini \cite{MalhotraS22} characterise projective MLN with only two variables and show that there are projective MLN that are not $\sigma$-determinate.
In particular, they show that in a binary vocabulary, every stochastic blocks model can be expressed by an MLN with two variables.
This shows that the direct equivalent of Proposition \ref{TPLP}, replacing PLP with MLN and determinate with $\sigma$-determinate, fails for MLN.\@
The notion of $\sigma$-projectivity allows us to pose this question in a revised form, left as a stimulus for further work:

Is an MLN $\sigma$-determinate if and only if it is $\sigma$-projective?

\section{Related work}
Our contribution immediately extends the recent work on projective families of distributions, which were introduced in \cite{JaegerS18}. 
A complete characterisation of projective families in terms of exchangeable arrays is provided in \cite{JaegerS20}, and a complete syntactic characterisaton of projective PLPs is presented in \cite{Weitkaemper21}. 
In Section \ref{Structured}, we extend their results to the practically essential case of structured input.

By enabling constant-time marginal inference and statistically consistent learning from samples, the study of projectivity lies in the wider field of lifted inference and learning \citep{VandenBroeckKNP21}. More precisely, projective families of distributions admit generalised lifted inference \citep{KhardonS21} 
In particular, Niepert and van den Broeck \cite{NiepertB21} study the connection between exchangeability and liftability. Since in light of Theorems 2 and 3 the study of projective families can equivalently be seen as the study of exchangeable distributions on infinite structures, it is enlightening to contrast our approach with the notion of exchangeability studied in \cite{NiepertB21}.
They consider exchangeability as invariance under permutations of the random variables encoded in the model, which is a much stronger assumption than invariance under permutations of the domain elements.
On the other hand, Niepert and Van den Broeck consider (partial) finite exchangeability rather than infinite exchangeability, with quite different behaviour from a probability-theoretic viewpoint \citep{Kallenberg05}. 

The results of Section \ref{Infinite} also provide a direct link between our work and previous work on statistical relational models for infinite domains. 
Among various other formalisms, previous work studied infinite models for RBNs \citep{Jaeger98a} and MLNs \citep{SinglaD07}. 
Our results in Section \ref{SigmaProj} help characterise $\sigma$-determinate MLNs that were introduced by Singla and Domingos as MLNs for infinite domains by providing $\sigma$-projectivity as a necessary condition for representability by a $\sigma$-determinate MLN.\@

In the restricted setting of random graphs rather than general relational structures, limits have been studied extensively in the theory of graphons and graph limits.
In particular, Corollary \ref{AHKinf} can be seen as a direct generalisation of \cite[Theorem 9.1]{DiaconisJ08} from graphs to the setting of general relational structures from \cite{JaegerS20}. 
Orbanz and Roy \cite{OrbanzR15} provide an overview of the field and its relationship to arrays such as the ones used in AHK models.
However, generalising graphon-oriented methods beyond simple graphs towards general relational structures is challenging, and even moving towards multi-relational graphs complicates the analysis considerably \citep{AlvaradoWR23}

\section{Conclusion}

By extending the concept of projectivity to structured input, we pave the way for applying projective families of distributions across the range of learning and reasoning tasks. 
We transfer the key results from projective families on unstructured input to structured input, including the motivating inference and learning properties \citep{JaegerS18}, the AHK representation \citep{JaegerS20} and the characterisation of projective PLPs \citep{Weitkaemper21}. 
We also gain some insight into possible applications, Corollary \ref{aritylimit} limiting the expressiveness for some common families of tasks. 
We then demonstrate the close connection between exchangeable distributions on infinite domains and projective families of distributions, which leads us to generic structures of vocabularies that extend this correspondence to structured input. 
Theorems \ref{ProjectiveFraisse} and \ref{ProjectiveInfinite} show how one can use projective families of distributions for models of potentially infinite streams of structured data, in which only an initial fragment is available for inspection at any given time. 
Finally, in Section \ref{SigmaProj} we apply the extension of projectivity to structured input to analyse projective families on unstructured input. 
This allows us to show that $\sigma$-determinate MLNs are $\sigma$-projective, which fundamentally distinguishes them from determinate PLPs.

\section*{Acknowledgments}

The research leading to this publication was supported by LMUexcellent, funded by the Federal
Ministry of Education and Research (BMBF) and the Free State of Bavaria under the Excellence Strategy of the Federal Government and the Länder. 

In addition, I would like to thank Kailin Sun for proofreading this paper, Nemi Pelgrom and Kilian R\"uckschlo{\ss} for lively discussion at the research group seminar, and the anonymous reviewers for their suggestions, which improved the manuscript. 

  \bibliographystyle{elsarticle-num} 
  \bibliography{Probabilisticlogicbib}
\end{document}